\documentclass{article}

\PassOptionsToPackage{numbers, compress}{natbib}

\usepackage[final]{ewrl_2025}

\usepackage[utf8]{inputenc} 
\usepackage[T1]{fontenc}    
\usepackage[hidelinks]{hyperref}       
\usepackage{url}            
\usepackage{booktabs}       
\usepackage{amsfonts}       
\usepackage{nicefrac}       
\usepackage{microtype}      
\usepackage{xcolor}         

\usepackage{amsmath}
\usepackage{amssymb}
\usepackage{amsthm}
\usepackage{graphicx}
\usepackage{subcaption}

\usepackage{hyperref}

\usepackage[capitalize,noabbrev]{cleveref}

\theoremstyle{plain}
\newtheorem{theorem}{Theorem}[section]

\newtheorem{lemma}[theorem]{Lemma}

\theoremstyle{definition}

\newtheorem{assumption}[theorem]{Assumption}
\theoremstyle{remark}

\usepackage{wrapfig}

\title{$K$-Level Policy Gradients for\\Multi-Agent Reinforcement Learning}

\author{%
  Aryaman Reddi$^{1,2}$\thanks{Correspondence to \texttt{<aryaman.reddi@tu-darmstadt.de>}}, Gabriele Tiboni$^{5}$, Jan Peters$^{1,2,3,4}$, Carlo D'Eramo$^{1,2,5}$ \\
  $^1$Department of Computer Science, TU Darmstadt, Germany \\
  $^2$Hessian Center for Artificial Intelligence (Hessian.ai), Germany \\
  $^3$German Research Center for AI (DFKI), Systems AI for Robot Learning, Germany \\
  $^4$Center for Cognitive Science, TU Darmstadt, Germany \\
  $^5$Center for Artificial Intelligence and Data Science, University of Würzburg, Germany \\
}

\begin{document}

\maketitle


\begin{abstract}
    Actor-critic algorithms for deep multi-agent reinforcement learning (MARL) typically employ a policy update that responds to the current strategies of other agents.
    While being straightforward, this approach does not account for the updates of other agents at the same update step, resulting in miscoordination.
    In this paper, we introduce the $K$-Level Policy Gradient (KPG), a method that recursively updates each agent against the updated policies of other agents, speeding up the discovery of effective coordinated policies.
    We theoretically prove that KPG with finite iterates achieves monotonic convergence to a local Nash equilibrium  under certain conditions.
    We provide principled implementations of KPG by applying it to the deep MARL algorithms MAPPO, MADDPG, and FACMAC. Empirically, we demonstrate superior performance over existing deep MARL algorithms in StarCraft II and multi-agent MuJoCo.
\end{abstract}

\section{Introduction}
\label{sec:introduction}

Deep multi-agent reinforcement learning (MARL) research has made strides towards solving practical problems such as cooperative robotics~\citep{ismail2018survey}, transportation management~\citep{haydari2020deep} and network traffic optimization~\citep{pi2024applications}. While deep RL algorithms have garnered impressive results in complex single-agent control problems~\citep{mnih2015human,tang2024deep}, multi-agent systems present unique challenges. Roadblocks in MARL research include exploding joint state-action spaces and non-stationarity due to concurrent learning~\citep{Li2009Multi-task,Barfuss2021Modeling}. 



Another challenge caused by simultaneous learning in MARL is a lack of mutual consistency, as each agent update does not account for the updates of other agents at the same update step, leading to miscoordination. To enhance the learning dynamics in MARL systems, we propose $K$-level Policy Gradients (KPG), a method which can be applied to any existing multi-agent policy gradient algorithm which can utilize a centralized policy gradient update. Most game-theoretic frameworks assume mutual consistency between agents; that is, each player's beliefs are consistent with what other players will actually do~\citep{robertson1936general}. In practical scenarios, agents must infer the strategies of others using modeling or recurrent reasoning. The cognitive hierarchies framework formalizes \textit{$k$-level thinking} as an iterative decision process whereby agents respond to the strategies they believe other agents will take, in accordance with other agent's updated strategies based on the same recursive reasoning: `I think that you think that I think...' ~\citep{camerer2004cognitive}. The prior distribution with which an agent acts is typically defined as recursion level $k{=}0$. MARL algorithms that update against the $k{=}0$ strategies of other agents can therefore be considered level $k{=}1$. KPG enables agents to update their policies recursively against the updated policies of other agents, allowing them to utilize higher orders of $k$-level reasoning. Aided with an enhanced notion of mutual consistency, multi-agent systems are therefore able to converge faster on coordination problems. The intuition behind this algorithm is inspired by human behavior: research on theory of mind suggests that humans often model the decision-making processes of other humans in order to successfully collaborate on complex tasks~\citep{gurney2021operationalizing, schaafsma2015deconstructing}.

Research in both on-policy and off-policy MARL methods have produced impressive results from the extension of single-agent RL methods. In this work, we demonstrate the effectiveness of centralized learning with KPG across several metrics in on-policy and off-policy settings. We first conduct a theoretical study of the algorithm in the idealized case, followed by a theorem that demonstrates the convergence of KPG with finite iterates to a local Nash equilibrium under reasonable conditions. We accompany our theoretical analysis with an illustrative example in a simple toy problem. Finally, we present a comparison of KPG against existing deep MARL algorithms in three complex multi-agent coordination environments: the multi-agent StarCraft II environment in JaxMARL~\citep{rutherford2023jaxmarl} (SMAX), the StarCraft II micromanagement benchmark in PyMARL~\citep{samvelyan2019starcraft} (SMAC), and multi-agent MuJoCo (MAMuJoCo)~\citep{peng2021facmac}. Note that SMAX and SMAC have different dynamics and should be considered separate baselines. In order to demonstrate the effectiveness of KPG in deep MARL, we practically implement it on top of the competitive MARL algorithms MAPPO~\citep{yu2103surprising}, FACMAC~\citep{peng2021facmac}, and MADDPG~\citep{lowe2017multi}, which we denote \textbf{$\boldsymbol{K}$-MAPPO}, \textbf{$\boldsymbol{K}$-FACMAC}, and \textbf{$\boldsymbol{K}$-MADDPG} respectively.

\section{Related works}
\paragraph{Policy gradient methods in multi agent reinforcement learning} Policy gradient methods in cooperative MARL often fall within the CTDE regime~\citep{kraemer2016multi} by conditioning on global state information and centralized actions during training. Algorithms such as MADDPG \citep{lowe2017multi} utilize a multi-agent deterministic policy gradient using a centralized critic to exhibit more robust behavior than decentralized DDPG agents. \citet{foerster2018counterfactual} similarly propose COMA, which uses a counterfactual baseline term to alleviate the noise in their centralized gradients, a common problem in cooperative MARL leading to difficulties in credit assignment. Concerned with credit assignment noise, \citet{du2019liir} implement a framework that learns a separate proxy reward in order to discriminatingly credit agents in multi-agent actor-critic methods. 
Certain actor-critic MARL algorithms such as FACMAC\citep{peng2021facmac} borrow from the rich literature on multi-agent value decomposition methods~\citep{sunehag2017value,son2019qtran, rashid2020weighted,rashid2020monotonic, wang2020qplex, pan2021softmax} which attempt to improve credit assignment by decomposing global value estimates into factored ones. 

\citet{yu2103surprising} conduct a comprehensive study examining the performance of MAPPO, a CTDE variant of PPO which conditions its advantage estimation on global state information. 
While the policy gradient methods mentioned have displayed impressive results, they suffer from instability caused by a lack of mutual consistency, since they all update agents without taking the simultaneous updates of other agents into account.  

LOLA~\citep{foerster2017learning} comes the closest to our approach methodologically, as it utilizes higher-order gradient terms to incorporate the future opponent policy into the learning update of each agent. This might be considered a $k{=}2$ level update. COLA~\citep{willi2022cola} addresses the high variance and difficult computation of higher-level LOLA iterates by minimizing a differentiable consistency measure, requiring only second-order derivatives. POLA~\citep{zhao2022proximal} builds on LOLA further, reinterpreting it as a proximal operator by penalizing divergence over policy behavior. This helps mitigate LOLA's sensitivity to parameterizations. While LOLA, COLA, and POLA are evaluated in the deep setting with two-player reciprocity-based games, their viability in more complex cooperative environments has not yet been sufficiently examined. We show in this work that the informed updates of KPG are more stable than POLA and can be successfully applied to achieve SOTA performance in complex environments. M-FOS~\citep{khan2023scaling} scales opponent shaping to general-sum games; while using a similar paradigm to LOLA, we do not benchmark against M-FOS as we consider its meta-game formulation out of the scope of this paper.
\vspace{-10pt}
\paragraph{$\boldsymbol{K}$-level reasoning in reinforcement learning}
$K$-level thinking has proven useful in several multi-agent opponent modeling scenarios. Notably, the cognitive hierarchies framework~\citep{camerer2004cognitive} has been combined extensively with deep RL techniques for the training of self-driving vehicles which must cooperate within a heterogeneous population of peer vehicles~\citep{li2016hierarchical,garzon2019game,bouton2020reinforcement, albaba2021driver,wang2022comprehensive, karimi2023level, dai2023game}.
~\citet{cui2021k} propose a synchronous hierarchical $k$-level reasoning method known as SyKLRBR which obtains SOTA performance in multi-agent Hanabi~\citep{bard2020hanabi}.

Finally, ~\citet{liu2022recursive} propose Level $k$ gradient play, a recursive reasoning algorithm which stabilizes GAN training~\citep{goodfellow2020generative}. We extend their findings on the Semi-Proximal Point Method in this work to N-player, general sum games.    

\section{Preliminaries}
We consider a multi-agent extension of a Markov decision process (MDP)~\citep{puterman2014markov} known as a Markov game~\citep{littman1994markov}, defined by a tuple $\mathcal{G}=\langle\mathcal{I},\mathcal{S},\mathcal{A},\mathcal{P},\mathcal{R},\gamma,N,\iota\rangle$. $\mathcal{I}\equiv\{1,\dots,N\}$ is the set of agents, $\mathcal{S}$ is the state space, $\mathcal{A} \equiv \times_{i\in \mathcal{I}}A_{i}$ is the joint action space of the agents. At each timestep, agent $i$ samples an action $a_{i}\in A_{i}$ from policy $\pi_{i}\left(a_{i}|s\right)$ parameterized by $\boldsymbol{\theta}_{i} \in \mathbb{R}^{d_{i}}$. The joint action of all agents $\boldsymbol{a}$ from the joint policy $\boldsymbol{\pi}(.|s)$ determines the next state according to the joint state transition function $\mathcal{P}\left( s'|s,\boldsymbol{a} \right)$. $\mathcal{R}\equiv\{R_{1},\dots,R_{N}\}$ are the set of agent reward functions. Each agent $i$ has a learning rate $\eta_{i}$ and receives rewards according to its reward function $R_{i}\left(s, \boldsymbol{a}, s' \right)$. $\gamma$ is the discount factor and $\iota$ is the initial state distribution. 
$\underset{max}{\lambda}(\boldsymbol{Z})$ and $\underset{min}{\lambda}(\boldsymbol{Z})$ refer to the maximum and minimum eigenvalues of a matrix $\boldsymbol{Z}$ respectively. Each agent aims to maximize its own discounted objective

\vspace{-10pt}
\begin{equation}
    J_{i}\left(\boldsymbol{\theta}_{i},\boldsymbol{\theta}_{-i}\right) = \underset{\tau\sim\ \iota, \pi_{i},\boldsymbol{\pi}_{-i},\mathcal{P}}{\mathbb{E}}\left[\sum_{t=0}^{\infty} \gamma^t r_{t,i}\right],
    \label{E:ma_objective}
\end{equation}

where $\boldsymbol{\pi}_{-i}$ and $\boldsymbol{\theta}_{-i}$ refer to the other agent policies and parameters respectively, $r_{t,i}$ is the reward agent $i$ receives at time $t$, and $\tau$ refers to the trajectories induced under the agent policies, $\iota$, and $\mathcal{P}$.

\section{$K$-level policy gradients}
In practice, $\nabla_{\boldsymbol{\theta}_{i}}J_{i}\left(\boldsymbol{\theta}_{i},\boldsymbol{\theta}_{-i}\right)$ can be estimated in a variety of ways, such as sample-based estimates of the policy gradient~\citep{lowe2017multi, lyu2021contrasting} or using the performance difference lemma~\citep{kakade2002approximately}. The optimization performed by agent $i$ is a response to the action distribution of the other agents before they have made an update. The update for the initial parameters $\boldsymbol{\theta}^{(0)}_{i}$ for agent $i$ results in a $k{=}1$ policy denoted $\boldsymbol{\theta}^{(1)}_{i}$, where the superscript denotes the degree of $k$-level thinking (we drop the superscript for $\boldsymbol{\theta}^{(0)}_{i}$ from now on for brevity when possible). We can write the $k{=}1$ policy gradient update as:

\vspace{-10pt}
\begin{align}
\label{eq:vanilla_pg_gradient}
\boldsymbol{\theta}_{i}^{(1)} &\gets \boldsymbol{\theta}_{i} + \eta_{i}\nabla_{\boldsymbol{\theta}_{i}}
J_{i}\left(\boldsymbol{\theta}_{i},\boldsymbol{\theta}_{-i} \right) \text{ , }\forall i\in\mathcal{I}.
\end{align}

\begin{figure}[] 
   \centering
   
   \includegraphics[width=\columnwidth]{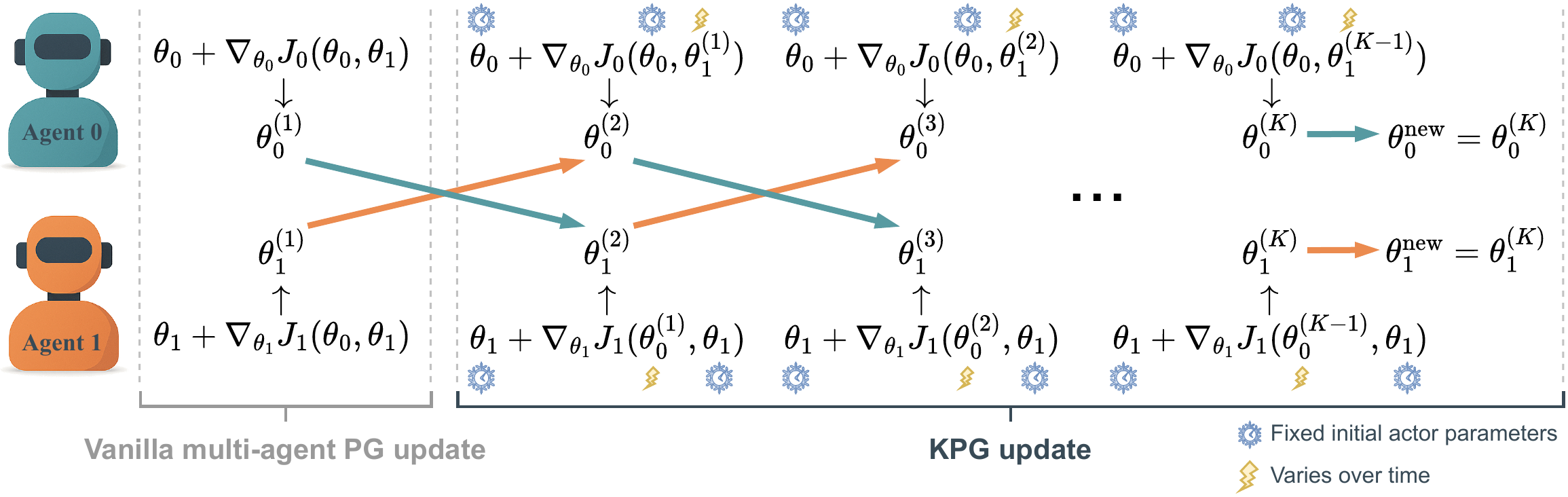}
   \caption{Schematic illustration of the $K$-level Policy Gradient (KPG) update for two agents.}
   \label{fig:kpg_illustration}
\end{figure}

The generalized update in \eqref{eq:vanilla_pg_gradient} is the standard one used in multi-agent policy gradient methods. The $k{=}1$ policy obtained in this way does not account for the fact that the other agent policies were also updated in the same step.  This update step can be taken once again from the \textit{initial} parameters of each agent while considering the $k{=}1$ policies of the other agents, allowing us to reach recursion level $k{=}2$:

\vspace{-10pt}
\begin{align}
\label{E:k1_update}
\boldsymbol{\theta}_{i}^{(2)} &\gets \boldsymbol{\theta}_{i} + \eta_{i}\nabla_{\boldsymbol{\theta}_{i}},
J_{i}\left(\boldsymbol{\theta}_{i},\boldsymbol{\theta}_{-i}^{(1)} \right) \text{ , }  \forall i\in\mathcal{I},
\end{align}

This procedure can be applied recursively \(K\) times, with each agent responding to the other agents at the previous $k$-level. Notably, the final $K$-level policy of each agent is still only \emph{one} gradient step away from the initial actor parameters, since each recursive update is applied to the same initial actor parameters \( \boldsymbol{\theta}\). \textbf{The $k$-level updates are not moving further in parameter space, but rather finding a gradient direction that is more consistent with the updates of the other agents.}
We show the full recursive reasoning procedure in Algorithm~\ref{A:KPG}, with a schematic illustration in Figure~\ref{fig:kpg_illustration} in the case of two agents. 

\begin{algorithm}
   \caption{$K$-Level Policy Gradients (KPG)}
   \label{A:KPG}
\begin{algorithmic}
   \STATE {\bfseries Input:} Initial actor parameters $\boldsymbol{\theta}_{i}$ \& actor learning rates $\eta_{i}$, $\forall  i\in\mathcal{I}$; recursive reasoning steps $K$
   \FOR{each update step}
   \FOR{$k=1$ {\bfseries to} $K$}
   \STATE $\boldsymbol{\theta}_{i}^{(k)} \leftarrow \boldsymbol{\theta}_{i} + \eta_{i}\nabla_{\boldsymbol{\theta}_{i}}J_{i}(\boldsymbol{\theta}_{i}, \boldsymbol{\theta}_{-i}^{(k-1)}), \; \forall i\in\mathcal{I}$ 
   \ENDFOR
   \STATE {\bfseries return} $\boldsymbol{\theta}_{i}^{(K)}$,  $\forall i\in\mathcal{I}$
   \ENDFOR
\end{algorithmic}
\end{algorithm}

\subsection{Theoretical analysis}
In this section, we present a theoretical study which shows that under certain conditions, KPG converges monotonically to a local Nash equilibrium, even with finite $K$ iterates. Firstly, we demonstrate how an unbiased, infinite application of Algorithm~\ref{A:KPG} leads to perfect anticipation of other agents' future strategies.

\begin{assumption}\label{ass:lipschitz}
\textit{The gradient $\nabla_{\boldsymbol{\theta_{i}}}
J_{i}\left(\boldsymbol{\theta}_{i},\boldsymbol{\theta}_{-i} \right)$ is $L_{i}$-Lipschitz with respect to $\boldsymbol{\theta}_{-i}$, i.e.}
\begin{align}\label{E:lipschitz}
    \| \nabla_{\boldsymbol{\theta_{i}}}
J_{i}\left(\boldsymbol{\theta}_{i,},\boldsymbol{\theta}_{-i,1} \right) - \nabla_{\boldsymbol{\theta_{i}}}
J_{i}\left(\boldsymbol{\theta}_{i,},\boldsymbol{\theta}_{-i,2} \right)\| \leq L_{i}\|\boldsymbol{\theta}_{-i,1} - \boldsymbol{\theta}_{-i,2}\|, \forall \boldsymbol{\theta}_{i}, \boldsymbol{\theta}_{-i},
\end{align}
\end{assumption}

where $\boldsymbol{\theta}_{-i,1}$ and $\boldsymbol{\theta}_{-i,2}$ are two arbitrary points in the joint parameter space of the other agents. We define the maximum objective function Lipschitzness $L := \underset{i}{\text{max}} \{L_{i}\}$ and the maximum agent learning rate $\eta := \underset{i}{\text{max}}\{\eta_{i}\}$. We define also the maximum objective function gradient across all agent parameters and objective functions
 $\nabla_{max} := \underset{i,\boldsymbol{\theta}_{i},\boldsymbol{\theta}_{-i}}{\text{max}}\|\nabla_{\boldsymbol{\theta}_{i}}J_{i}(\boldsymbol{\theta}_{i},\boldsymbol{\theta}_{-i})\|$.

\begin{theorem}\label{th:cauchy}
Suppose Assumption~\ref{ass:lipschitz} holds. We define the combined agent parameter vector $\boldsymbol{\theta} := \left[\boldsymbol{\theta}_{1}, ...,\boldsymbol{\theta}_{n}\right]^T \in \mathbb{R}^{\sum_{i} d_{i}}$. Let $\boldsymbol{\theta}^{(k)}$ be the combined agent parameter vector after $k$ steps of KPG reasoning. 
For $k$-level KPG, consecutive update steps are bounded: 
\begin{equation}
    \|\boldsymbol{\theta}^{(k)} - \boldsymbol{\theta}^{(k-1)}\| \leq \eta (\eta L)^{k-1}n(n-1)^{k-1}\nabla_{max}.
\end{equation}

Assume the maximum learning rate $\eta$ satisfies $\eta < \frac{1}{L(n-1)}$. Then, the sequence $\{\boldsymbol{\theta}^{(k)}\}_{k{=}0}^{\infty}$ is a convergent sequence. Since $\boldsymbol{\theta}$ exists in a complete subspace of $\mathbb{R}^{\sum_{i}d{i}}$, the convergent sequence $\{\boldsymbol{\theta}^{(k)}\}_{k=0}^{\infty}$ is Cauchy, i.e.,

\vspace{-10pt}
\begin{equation}
    \exists C\in\mathbb{N} : \forall \epsilon>0, (a>b>C \implies\|\boldsymbol{\theta}^{(a)} - \boldsymbol{\theta}^{(b)}\| < \epsilon).
\end{equation}

Since every Cauchy sequence has a limit, we denote the limit of $\{\boldsymbol{\theta}^{(k)}\}_{k{=}0}^{\infty}$ as $\underset{k\rightarrow\infty}{lim}\boldsymbol{\theta}^{(k)} = \boldsymbol{\theta}^{(\infty)}$.
\end{theorem}

According to Theorem~\ref{th:cauchy}, applying the update in Algorithm~\ref{A:KPG} with $k{=}\infty$ defines the following implicit algorithm: 

\vspace{-10pt}
\begin{equation}
    \boldsymbol{\theta}_{i}^{(\infty)} \leftarrow  \boldsymbol{\theta}_{i} + \eta_{i}\nabla_{\boldsymbol{\theta}_{i}}
J_{i}\left(\boldsymbol{\theta}_{i},\boldsymbol{\theta}_{-i}^{(\infty)}\right), \forall i\in\mathcal{I},
\end{equation}

which we denote the Generalized Semi-Proximal Point Method (GSPPM). The implication of the GSPPM is that the update of each agent responds exactly to the updated strategies of the other agents, maintaining mutual consistency.~\citet{liu2022recursive} similarly establish the SPPM in $2$-player minimax games, and we extend their findings to $N$-player general sum games in this work. 

Following from Theorem~\ref{th:cauchy}, we show that the convergence of GSPPM iterates in a non-convex non-concave strategy space can be analyzed via the game Jacobian around a local stationary point:
\begin{theorem}\label{th:GSPPM}
Let $\boldsymbol{\theta}^{*}$ be a stationary point in an N-player general sum game. This stationary point is a local Nash equilibrium, i.e. a point at which no agent's objective function has a non-zero gradient under a unilateral change in policy. Let $\boldsymbol{\eta}$ be a block matrix of the agent learning rates $\eta_{i}$.  Let the components of the Hessian of each objective function at $\boldsymbol{\theta}^{*}$ be denoted

\vspace{-10pt}
$$
    \begin{pmatrix}
    \nabla_{\boldsymbol{\theta}_{i}\boldsymbol{\theta}_{i}}^{2}J_{i}(\boldsymbol{\theta}_{i}^{*}\boldsymbol{\theta}_{-i}^{*})   & \nabla_{\boldsymbol{\theta}_{i}\boldsymbol{\theta}_{-i}}^{2}J_{i}(\boldsymbol{\theta}_{i}^{*}\boldsymbol{\theta}_{-i}^{*}) \\
    \nabla_{\boldsymbol{\theta}_{-i}\boldsymbol{\theta}_{i}}^{2}J_{i}(\boldsymbol{\theta}_{i}^{*}\boldsymbol{\theta}_{-i}^{*})   & \nabla_{\boldsymbol{\theta}_{-i}\boldsymbol{\theta}_{-i}}^{2}J_{i}(\boldsymbol{\theta}_{i}^{*}\boldsymbol{\theta}_{-i}^{*})
\end{pmatrix}
 = \begin{pmatrix}
 \boldsymbol{A}_{i} & \boldsymbol{B}_{i} \\
 \boldsymbol{B}_{i}^{\top} & \boldsymbol{C}_{i}
 \end{pmatrix}.
$$

Furthermore, let $\boldsymbol{A}$ be the diagonal block matrix of all $\boldsymbol{A}_{i}$ matrices, and $\boldsymbol{B}$ be the diagonal block matrix of all $\boldsymbol{B}_{i}$ matrices:

\vspace{-10pt}
$$
\boldsymbol{A} = \operatorname{diag}(\boldsymbol{A}_1, \dots, \boldsymbol{A}_n), \quad
\boldsymbol{B} = \operatorname{diag}(\boldsymbol{B}_1, \dots, \boldsymbol{B}_n)
$$

Let $\boldsymbol{D}$ be a complement-selection matrix for each set of agent parameters $\boldsymbol{\theta}_{i}$ such that

\vspace{-10pt}
$$
\boldsymbol{D} \boldsymbol{\theta} = \left[\boldsymbol{\theta}_{-1}, ...,\boldsymbol{\theta}_{-n}\right]^\top
$$

Suppose $\eta < \frac{1}{L(n-1)}$ such that the GSPPM iterates $\{\boldsymbol{\theta}_{t}^{(k)}\}_{k{=}0}^{\infty}$ form a Cauchy sequence. Then, there exists a neighborhood $\mathcal{U} \in \mathbb{R}^{\sum_{i}d_{i}}$ around $\boldsymbol{\theta}^{*}$ such that if GSPPM starts in $\mathcal{U}$, the iterates $\{\boldsymbol{\theta}_{t}^{(k)}\}_{k=0}^{\infty}$ satisfy:

\vspace{-10pt}
\begin{equation}
    \|\boldsymbol{\theta}^{(\infty)} - \boldsymbol{\theta}^{*} \| \leq \frac{\underset{max}{\lambda}(I + \boldsymbol{\eta}\boldsymbol{A})^{2}}{\underset{min}{\lambda}(I - \boldsymbol{\eta}\boldsymbol{B}\boldsymbol{D})^{2}}\|\boldsymbol{\theta} - \boldsymbol{\theta}^{*} \|.
\end{equation}

Moreover, for any $\boldsymbol{\eta}$ satisfying

\vspace{-10pt}
\begin{equation}
\frac{\underset{max}{\lambda}(I + \boldsymbol{\eta}\boldsymbol{A})^{2}}{\underset{min}{\lambda}(I - \boldsymbol{\eta}\boldsymbol{B}\boldsymbol{D})^{2}} < 1,
\end{equation}

the iterates converge asymptotically to $\boldsymbol{\theta}^{*}$.

\end{theorem}

Hence, GSPPM iterates reach an $\epsilon$-Nash equilibrium, i.e., an arbitrarily close $\epsilon$-bound around the stationary point~\citep{fudenberg1998theory}, under the conditions of Theorems~\ref{th:cauchy} and~\ref{th:GSPPM}.

\begin{theorem}\label{th:KPG_k_convergence}
Suppose the conditions of Theorem~\ref{th:GSPPM} apply. Then, the distance of finite iterates $\{\boldsymbol{\theta}_{t}^{(k)}\}$ generated by KPG satisfy 

\vspace{-10pt}
\begin{align}
    \|\boldsymbol{\theta}^{(k)} - \boldsymbol{\theta}^{*} \|^2  & \leq  \left(\underset{max}{\lambda}(I + \boldsymbol{\eta} \boldsymbol{A})^2 
    + 2\underset{max}{\lambda}(I+\boldsymbol{\eta} \boldsymbol{A})\underset{max}{\lambda}(\boldsymbol{\eta} \boldsymbol{B}\boldsymbol{D}) \right)\|\boldsymbol{\theta} - \boldsymbol{\theta}^{*} \|^2 \\
    &+ 2\underset{max}{\lambda}(I+\boldsymbol{\eta} \boldsymbol{A})\underset{max}{\lambda}(\boldsymbol{\eta} \boldsymbol{B}\boldsymbol{D})\left(\|\boldsymbol{\theta} - \boldsymbol{\theta}^{*} \|\nabla_{max} \right) \\
    & + \underset{max}{\lambda}(\boldsymbol{\eta} \boldsymbol{B}\boldsymbol{D})^2 \|\boldsymbol{\theta}^{(k-1)} - \boldsymbol{\theta}^{*} \|^2.
\end{align}
    
Moreover, for any $\boldsymbol{\eta}$ satisfying

\vspace{-10pt}
\begin{equation}
    \underset{max}{\lambda}(\boldsymbol{\eta} \boldsymbol{B}\boldsymbol{D})^2 < 1,
\end{equation}
the KPG iterates converge asymptotically to $\boldsymbol{\theta}^{*}$. 
\end{theorem}

Thus, KPG iterates with finite $k$ reach an $\epsilon$-Nash equilibrium under the conditions of Theorems~\ref{th:cauchy},~\ref{th:GSPPM} and~\ref{th:KPG_k_convergence}.

\subsection{Illustrative example}
\label{sec:illustrative_example}

\begin{figure}[t] 
    \centering
    \begin{subfigure}[t]{0.49\textwidth}
        \centering
        \includegraphics[width=0.8\textwidth]{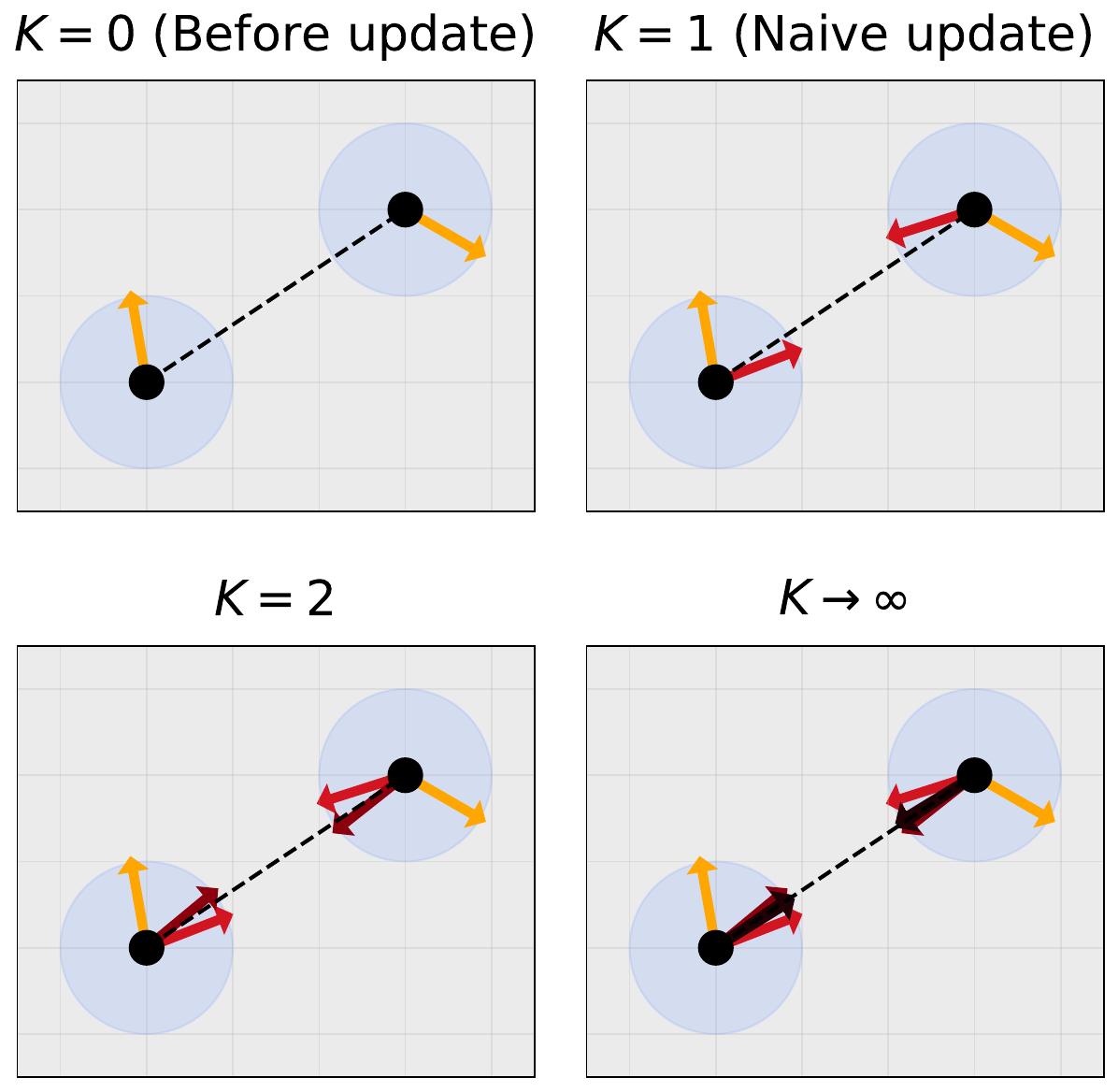}
        \caption{Agents taking continuous actions with increasing $k$-levels (darkening arrows) converging on the optimal actions (dashed line) within a single update step.}
        \label{fig:toy-agents}
    \end{subfigure}
    \hfill
    \begin{subfigure}[t]{0.49\textwidth}
        \centering
        \includegraphics[width=0.8\textwidth]{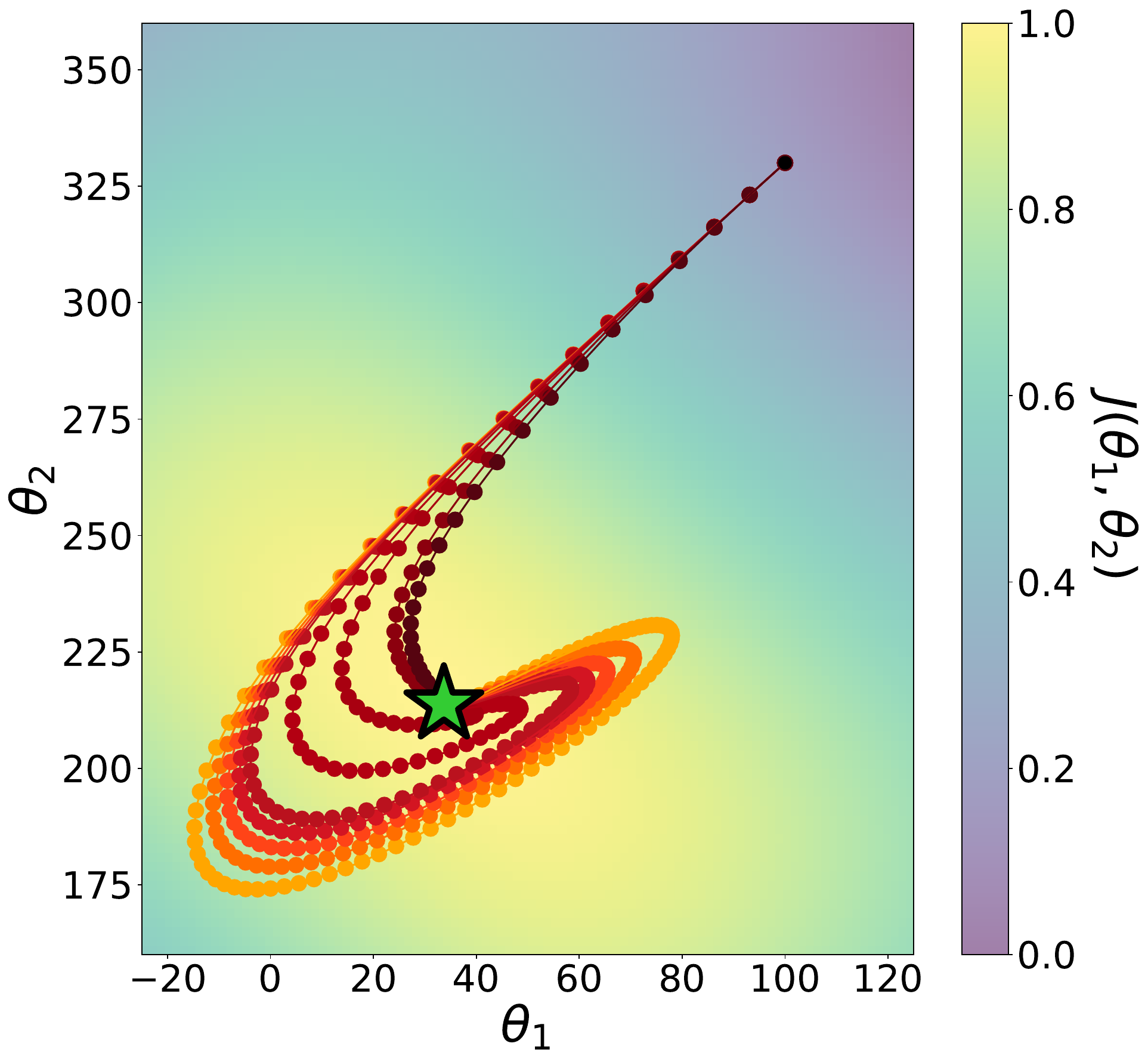}
        \caption{Gradient ascent to the optimal parameters (green star) with increasing $k$-levels (darker colors for higher $k$ values).}
        \label{fig:toy-heatmap}
    \end{subfigure}
    \caption{An illustrative continuous cooperative game with two point agents using $k$-level policy gradient ascent.}
    \label{fig:toy}
\end{figure}

We illustrate the effectiveness of KPG by studying a simple continuous cooperative game with two point agents taking continuous actions in a 2D space. The agents have one parameter each and produce a one-dimensional action (the angle of their next move).

The highest reward is achieved when the chosen direction of each agent points towards the \emph{future} location of the other agent. Assuming the agents move kinematically and have no momentum, this game has a known optimal solution: moving towards each other in a straight line. Therefore, the two agents should cooperate to meet as quickly as possible (see Appendix~\ref{appendix:illustrative_example} for details). However, the naive policy update for this game is for each agent to choose its next action to intercept with the \textit{previous} action of the other agent. The top two figures of Figure~\ref{fig:toy-agents} illustrate this problem: each agent's new action (red arrows) points towards the destination of the other agent under the other agent's old policy (yellow arrows). Hence, the naive update leads to a lack of mutual consistency since each agent does not consider the other agent's update.

\begin{wrapfigure}{r}{0.3\textwidth}
    \includegraphics[width=0.3\columnwidth]{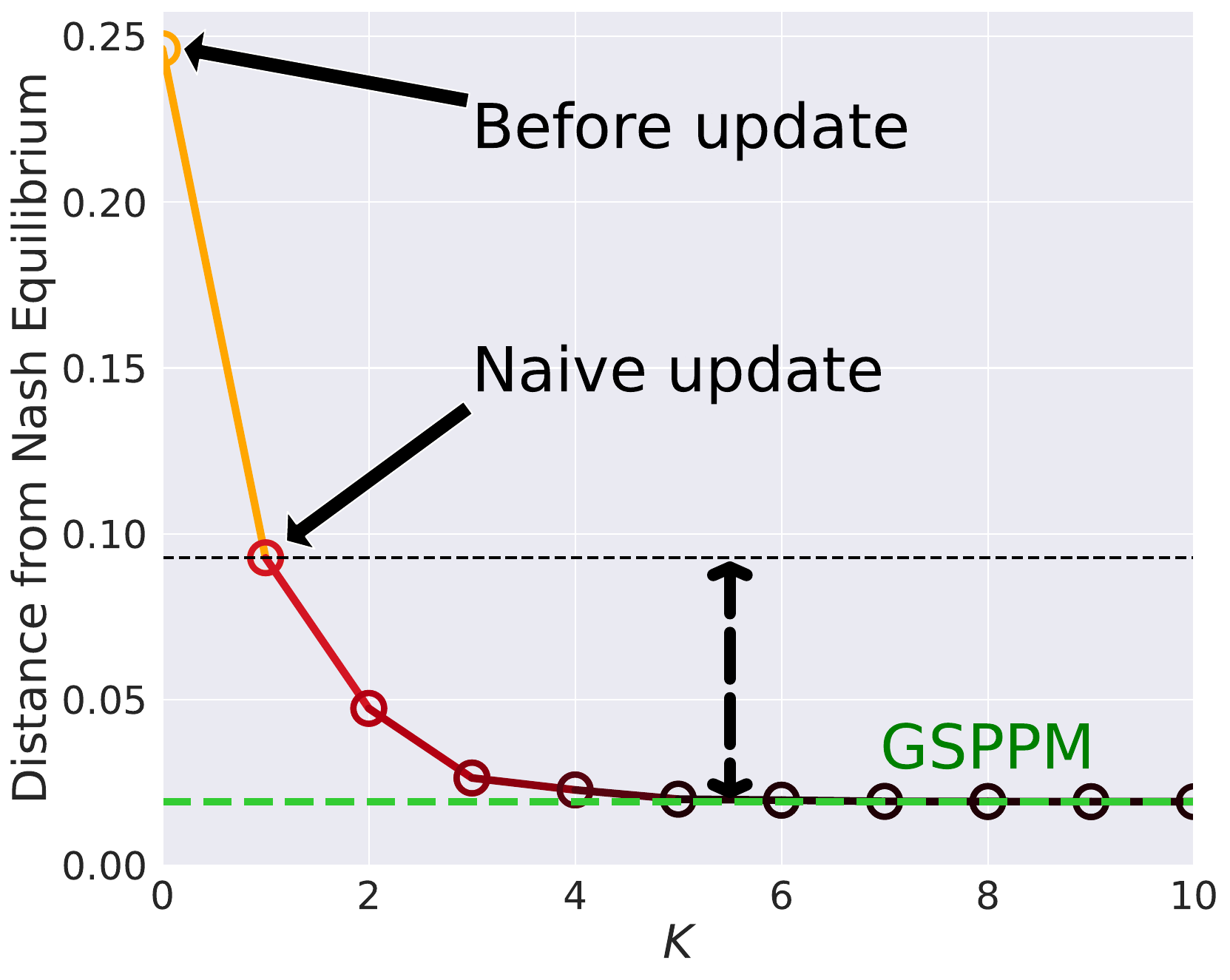}    
    \caption{Monotonic convergence to GSPPM for one update step with KPG.}
    \label{fig:toy-gsppm}
\end{wrapfigure}

Figure~\ref{fig:toy-agents} bottom left shows the policy update after $2$ levels of recursion: now the agents update to intercept the other agent \textit{after} the other agent's naive policy update, resulting in better coordination. As the number of recursions increases, the policies converge on an $\epsilon$-Nash bound of the optimal solution.


Figure~\ref{fig:toy-heatmap} shows the progression of the agent parameters with gradient ascent and momentum; we use the true gradient and objective function $J\left(\boldsymbol{\theta}_1, \boldsymbol{\theta}_2\right)$. The benefits of KPG recursion are evinced by the fact that increasing $k$-levels (darker points) exhibit monotonic convergence to the optimal parameters $\boldsymbol{\theta}_{1}^*,\boldsymbol{\theta}_{2}^*$ at each update step, as supported by Theorem~\ref{th:KPG_k_convergence}.
Figure~\ref{fig:toy-gsppm} shows the distance from the Nash equilibrium for one update step with KPG near the stationary point in Figure~\ref{fig:toy-heatmap}.

\subsection{Algorithmic implementation of the $k$-level policy gradient}
The KPG formulation is based on the recursive computation of the $k$-level policy gradient $\nabla_{\boldsymbol{\theta}_i}J(\boldsymbol{\theta}_i, \boldsymbol{\theta}_{-i}^{(k)})$.
However, unlike our illustrative example in Sec.~\ref{sec:illustrative_example}, this quantity can generally only be estimated, e.g.~through environment interactions.
We here propose two implementations of KPG on top of existing algorithms that bridge the gap between its theoretical formulation and its application for deep RL problems. In particular, we show how the $k$-level policy gradient can be estimated using \emph{only} the data collected by the initial joint policy $\boldsymbol{\pi}^{(0)}$ before the update, hence avoiding the need for additional environment interactions during the $k$-level reasoning process.

\noindent\textbf{K-MAPPO (on-policy).} We integrate KPG into the state-of-the-art algorithm MAPPO~\citep{yu2103surprising} by building on the surrogate objective provided by PPO~\citep{schulman2017proximal}. MAPPO utilizes a surrogate loss with state-conditioned advantage estimation and an IS ratio, $r_{s,\boldsymbol{a}}(\boldsymbol{\theta}_{i}, \boldsymbol{\theta}_{-i}) = \frac{\pi_{i}(a_{i}|s)}{\pi^{(0)}_{i}(a_{i}|s)}$ (where $\pi_{i}$ is the policy being updated), which does not consider the update of the other agents. We extend this ratio to include the updated policies:
\begin{equation}\label{eq:kmappo_is_ratio}
    r_{s,\boldsymbol{a}}(\boldsymbol{\theta}_i, \boldsymbol{\theta}_{-i}^{(k)}) = \frac{\pi_i(a_i|s) \cdot \boldsymbol{\pi}_{-i}^{(k)}(\boldsymbol{a}_{-i}|s)}{\boldsymbol{\pi}^{(0)}(\boldsymbol{a}|s)} = \frac{\pi_i(a_i|s)}{\pi_i^{(0)}(a_i|s)} \cdot \frac{\boldsymbol{\pi}_{-i}^{(k)}(\boldsymbol{a}_{-i}|s)}{\boldsymbol{\pi}_{-i}^{(0)}(\boldsymbol{a}_{-i}|s)}.
 \end{equation}
In turn, this allows us to estimate the $k$-level policy gradient for each agent $i$ as:
\begin{equation}\label{eq:kmappo_surrogate_objective}
    \begin{split}
    &\nabla_{\boldsymbol{\theta}_i}J_{i}(\boldsymbol{\theta}_i, \boldsymbol{\theta}_{-i}^{(k)}) \approx \
     \nabla_{\boldsymbol{\theta}_i}\mathcal{L}_{i}^{\text{K-MAPPO}}(\boldsymbol{\theta}_i, \boldsymbol{\theta}_{-i}^{(k)}) = \\
   &= \mathbb{E}_{(s,\boldsymbol{a}) \sim \boldsymbol{\pi}^{(0)}} \left[\nabla_{\boldsymbol{\theta}_i} \min\left(r_{s,\boldsymbol{a}}(\boldsymbol{\theta}_i, \boldsymbol{\theta}_{-i}^{(k)}), \text{clip}(r_{s,\boldsymbol{a}}(\boldsymbol{\theta}_i, \boldsymbol{\theta}_{-i}^{(k)}), 1-\epsilon, 1+\epsilon)\right)\cdot A^{\boldsymbol{\pi}^{(0)}}_{i}(s,\boldsymbol{a}) \right],
    \end{split}
\end{equation}
where $A^{\boldsymbol{\pi}^{(0)}}_{i}(s,\boldsymbol{a})$ is the advantage function of agent $i$ under the $0$-level policies. Intuitively, the $k$-level IS ratio in Eq.~\ref{eq:kmappo_is_ratio} can be seen as weighting the original surrogate loss; if the joint probability ratio of the other agent updates exceeds 1, it increases the magnitude of the gradient that would have been taken without considering their updates, and vice-versa.  A comprehensive derivation of the K-MAPPO surrogate objective is available in the Appendix.

\noindent\textbf{K-MADDPG \& K-FACMAC (off-policy).} In the case of off-policy approaches such as MADDPG~\citep{lowe2017multi}, we can estimate the $k$-level policy gradient by leveraging an explicitly learned centralized Q-function.
Analogously to how MADDPG relies on the gradient of the learned Q-function to execute a policy improvement step, we propose the following gradient estimate:
\begin{equation}
    \nabla_{\boldsymbol{\theta}_i} J_i(\boldsymbol{\theta}_i, \boldsymbol{\theta}_{-i}^{(k)}) \approx \
    \nabla_{\boldsymbol{\theta}_i}\mathcal{L}_{i}^{\text{K-MADDPG}}(\boldsymbol{\theta}_i, \boldsymbol{\theta}_{-i}^{(k)}) 
    =\underset{\substack{
        s \sim \mathcal{D}\\
        a_i, \boldsymbol{a}_{-i} \sim \pi_i, \boldsymbol{\pi}_{-i}^{k}}}{\mathbb{E}} \left[ \nabla_{\boldsymbol{\theta}_i} Q_i^{\pi^{(0)}_{i}}(s, a_i, \boldsymbol{a}_{-i}) \right],
\end{equation}
where $\mathcal{D}$ is a buffer of stored transitions and $Q_{i}^{\boldsymbol{\pi}^{(0)}}$ is agent $i$'s action value function under the joint $k{=}0$ level policy.
In other words, we adjust the policy improvement step at each $k$-level reasoning step by resampling the actions of the other agents from the newly updated policies $\boldsymbol{\pi}_{-i}^{(k)}$, whereas $\pi_{i}=\pi_{i}^{(0)}$ remains fixed. We also provide a theoretical motivation of this approximation in the Appendix.
Note that the same principle can be applied to more sophisticated off-policy algorithms such as FACMAC.

\section{Experimental results}

\begin{figure*}[h] 
    \centering
    \begin{subfigure}[t]{0.24\textwidth}
        \centering
    \includegraphics[width=\textwidth]{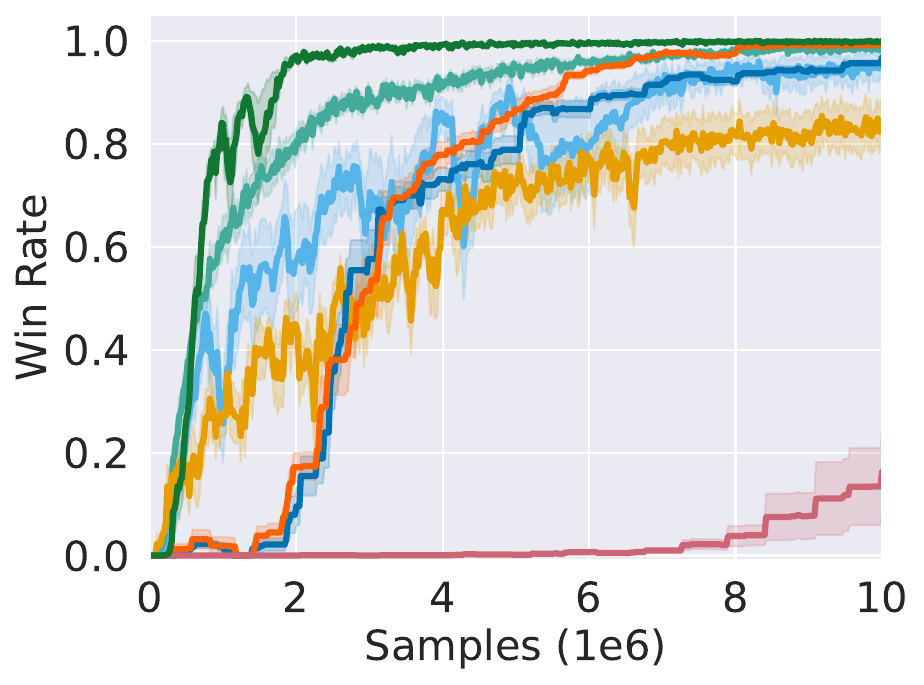}
        \caption{2s3z}
        \label{fig:smax-2s3z}
    \end{subfigure}
    \begin{subfigure}[t]{0.24\textwidth}
        \centering
        \includegraphics[width=\textwidth]{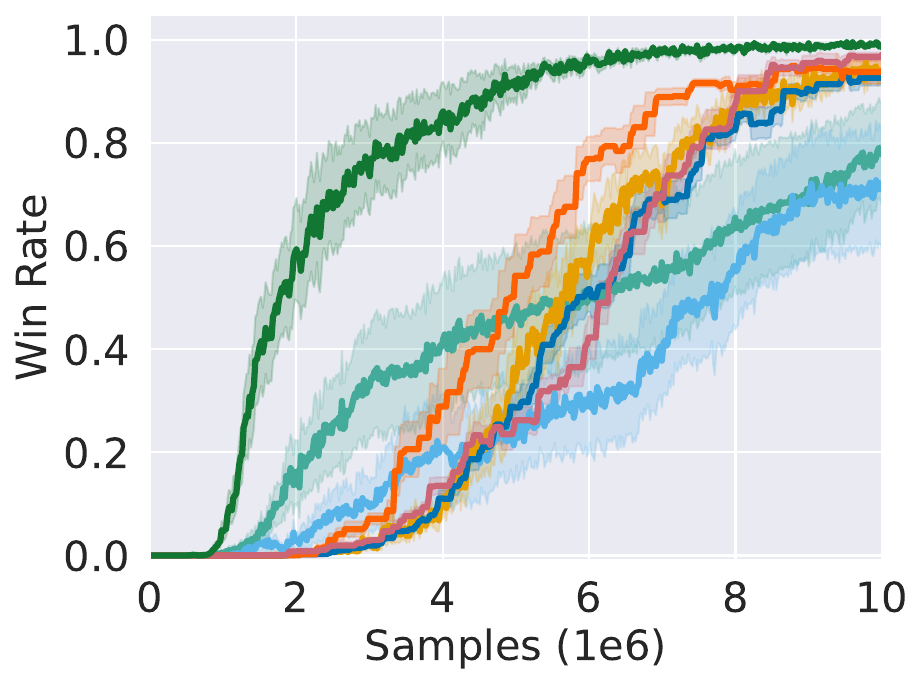}
        \caption{3s\_vs\_5z}
        \label{fig:smax-3svs5z}
    \end{subfigure}
    \begin{subfigure}[t]{0.24\textwidth}
        \centering
        \includegraphics[width=\textwidth]{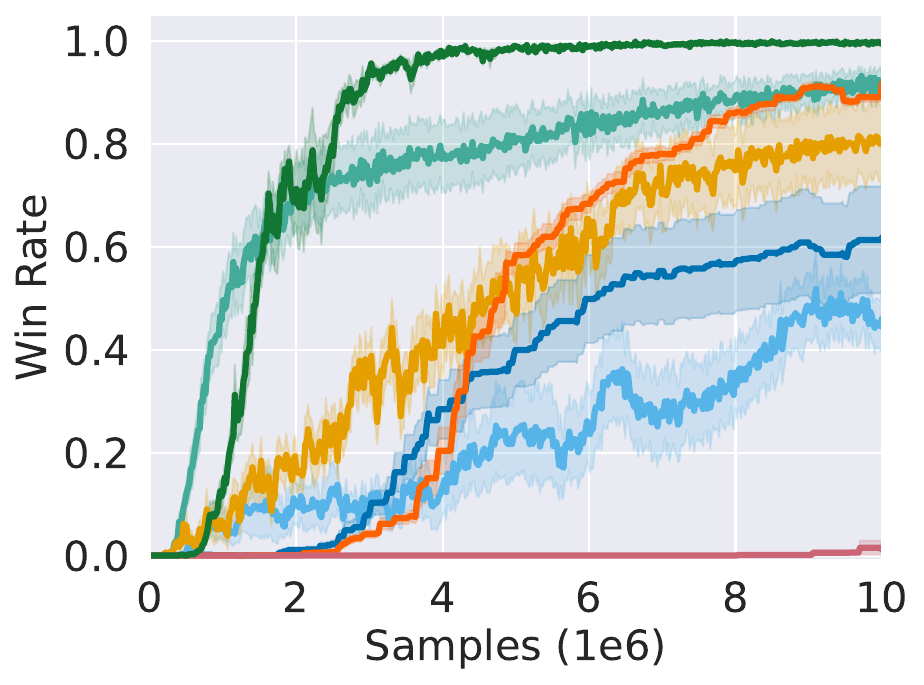}
        \caption{3s5z}
        \label{fig:smax-3s5z}
    \end{subfigure}
    \begin{subfigure}[t]{0.24\textwidth}
        \centering
        \includegraphics[width=\textwidth]{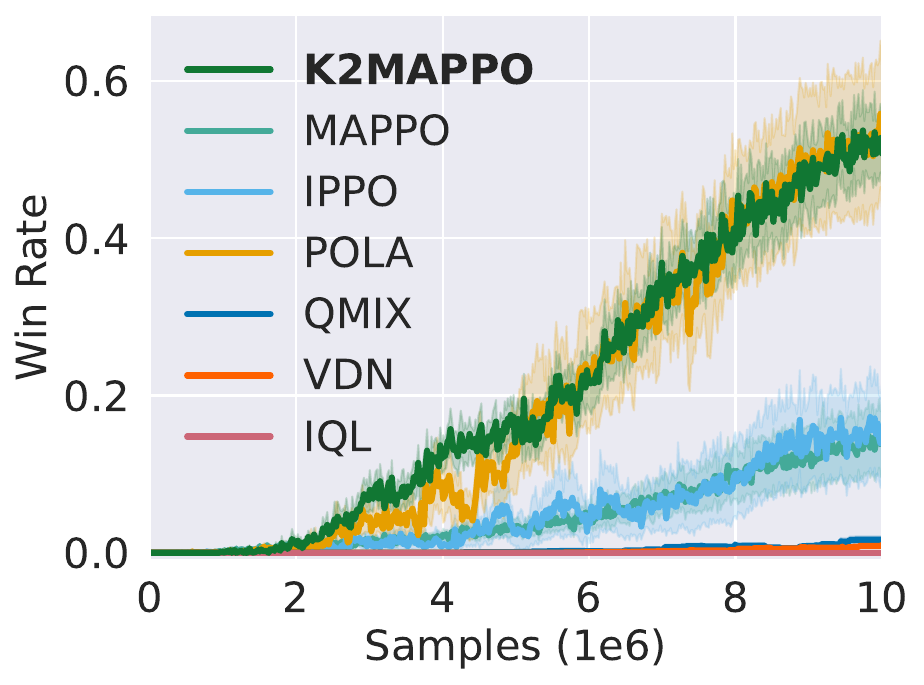}
        \caption{3s5z\_vs\_3s6z}
        \label{fig:smax-3s5z_vs_3s6z}
    \end{subfigure}

    \vspace{0.2cm}
    
    \begin{subfigure}[t]{0.24\textwidth}
        \centering
    \includegraphics[width=\textwidth]{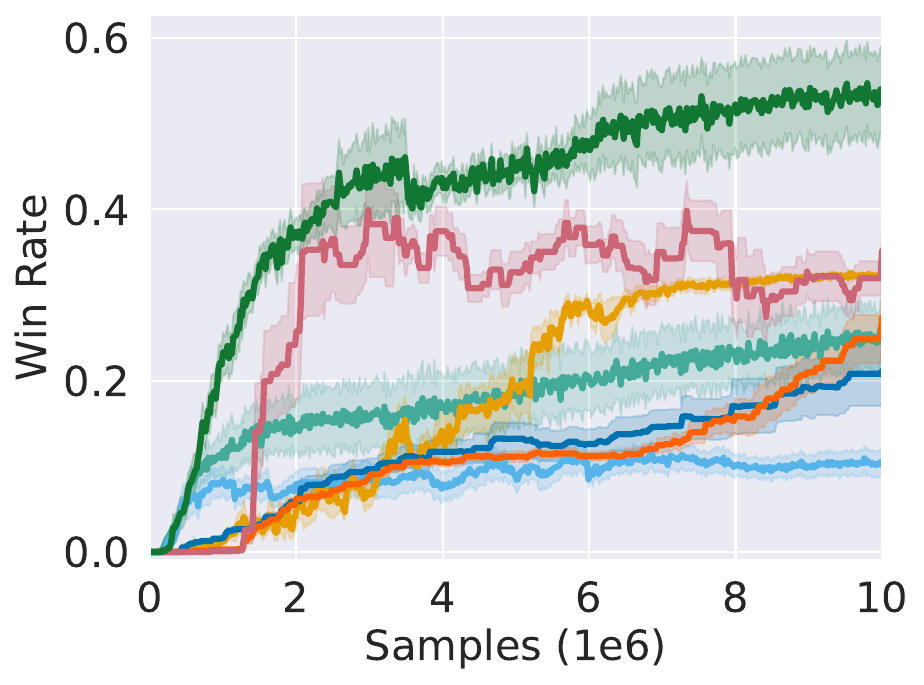}
        \caption{10m\_vs\_11m}
        \label{fig:smax-10m_vs_11m}
    \end{subfigure}
    \begin{subfigure}[t]{0.24\textwidth}
        \centering
        \includegraphics[width=\textwidth]{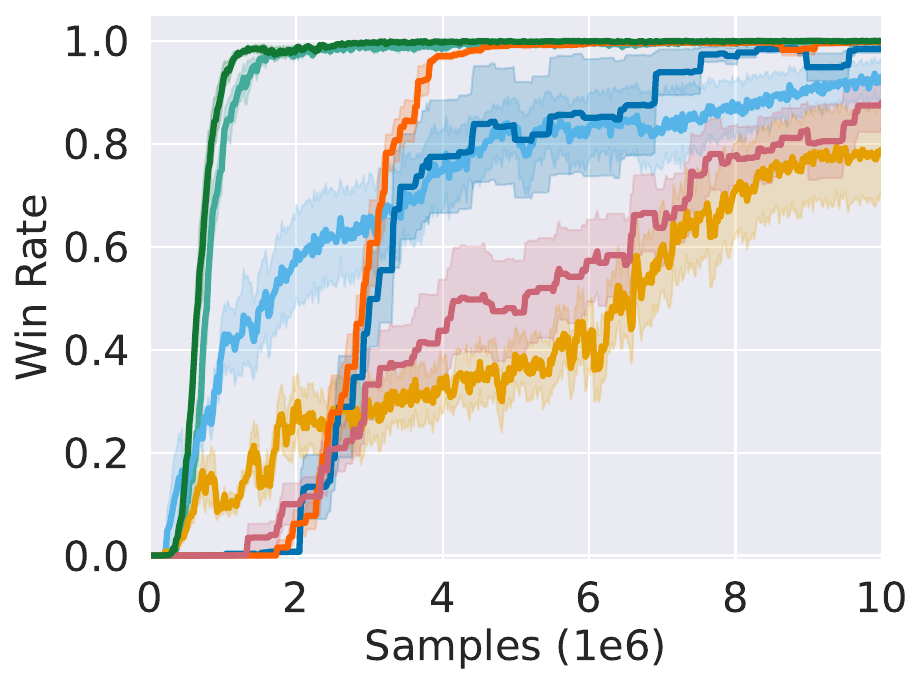}
        \caption{6h\_vs\_8z}
        \label{fig:smax-6h_vs_8z}
    \end{subfigure}
    \begin{subfigure}[t]{0.24\textwidth}
        \centering
        \includegraphics[width=\textwidth]{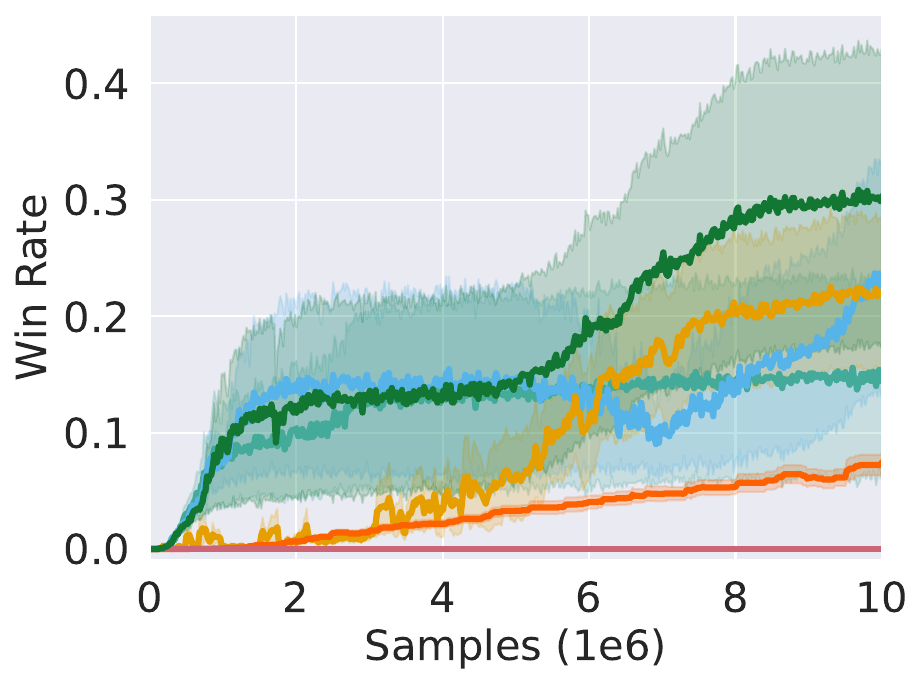}
        \caption{5m\_vs\_6m}
        \label{fig:smax-5m_vs_6m}
    \end{subfigure}
    \begin{subfigure}[t]{0.24\textwidth}
        \centering
        \includegraphics[width=\textwidth]{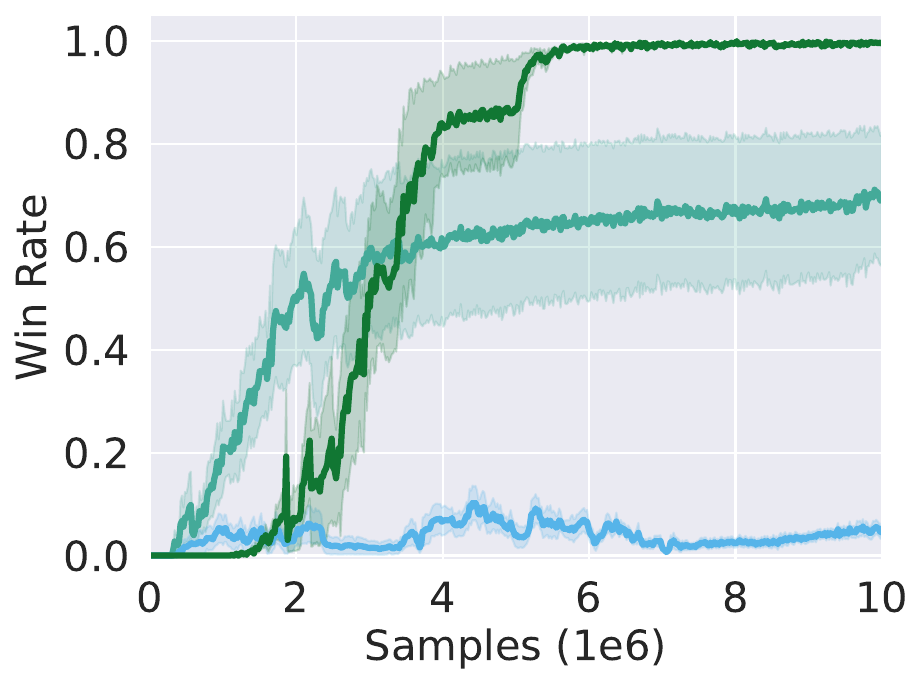}
        \caption{27m\_vs\_30m}
        \label{fig:smax-27m_vs_30m}
    \end{subfigure}

    \vspace{0.2cm}
    
    \begin{subfigure}[t]{0.24\textwidth}
        \centering
    \includegraphics[width=\textwidth]{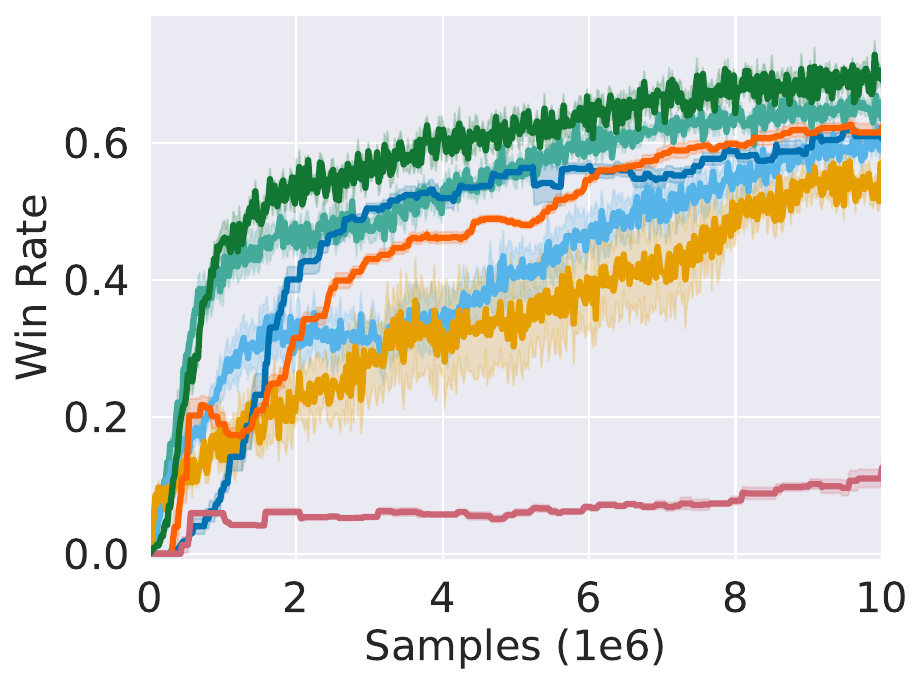}
        \caption{smacv2\_5\_units}
        \label{fig:smax-smacv2_5_units}
    \end{subfigure}
    \begin{subfigure}[t]{0.24\textwidth}
        \centering
        \includegraphics[width=\textwidth]{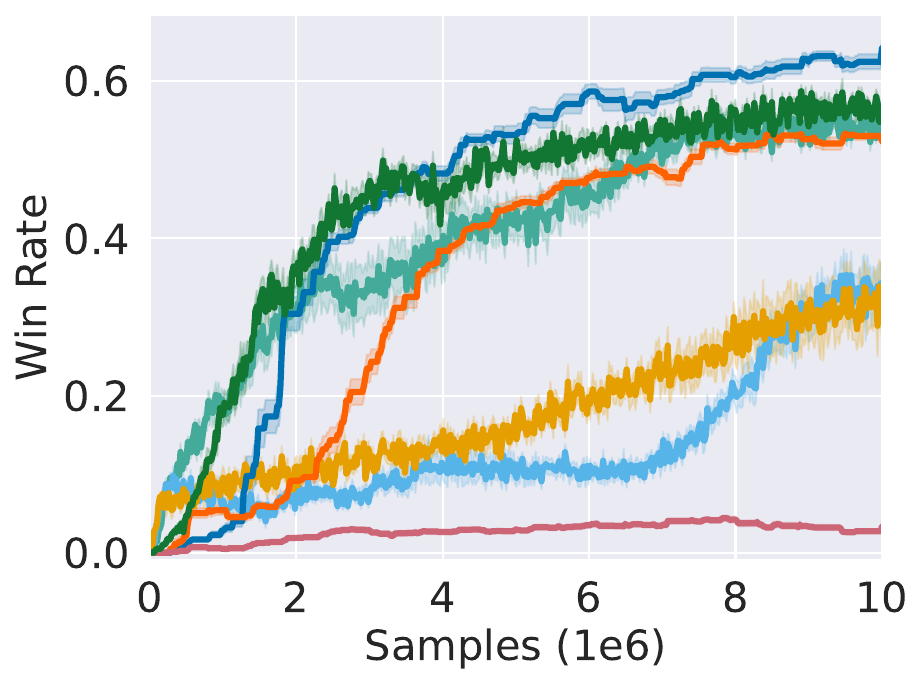}
        \caption{smacv2\_10\_units}
        \label{fig:smax-smacv2_10_units}
    \end{subfigure}
    \begin{subfigure}[t]{0.24\textwidth}
        \centering
        \includegraphics[width=\textwidth]{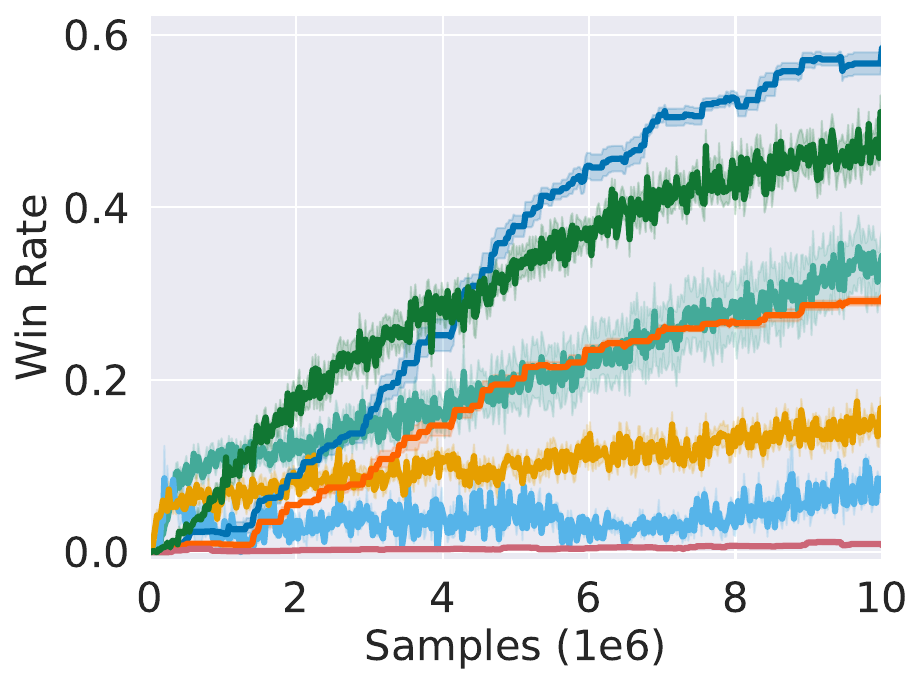}
        \caption{smacv2\_20\_units}
        \label{fig:smax-smacv2_20_units}
    \end{subfigure}
    \caption{Mean win rate and standard error of $K2$-MAPPO and baselines on SMAX maps across 10 seeds. Note that 27m\_vs\_30m is very large and could only be benchmarked with PPO-based methods due to computational constraints.}
    \label{fig:smax}
\end{figure*}

\begin{figure*}[t] 
    \centering
    \begin{subfigure}[t]{0.24\textwidth}
        \centering
    \includegraphics[width=\textwidth]{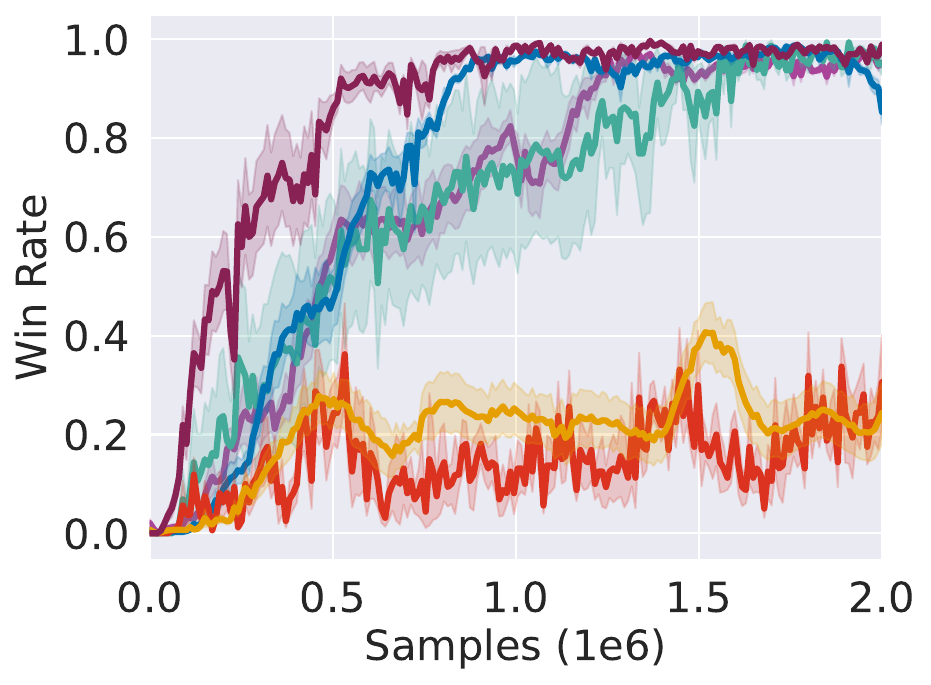}
        \caption{MMM}
        \label{fig:smac-MMM}
    \end{subfigure}
    \begin{subfigure}[t]{0.24\textwidth}
        \centering
        \includegraphics[width=\textwidth]{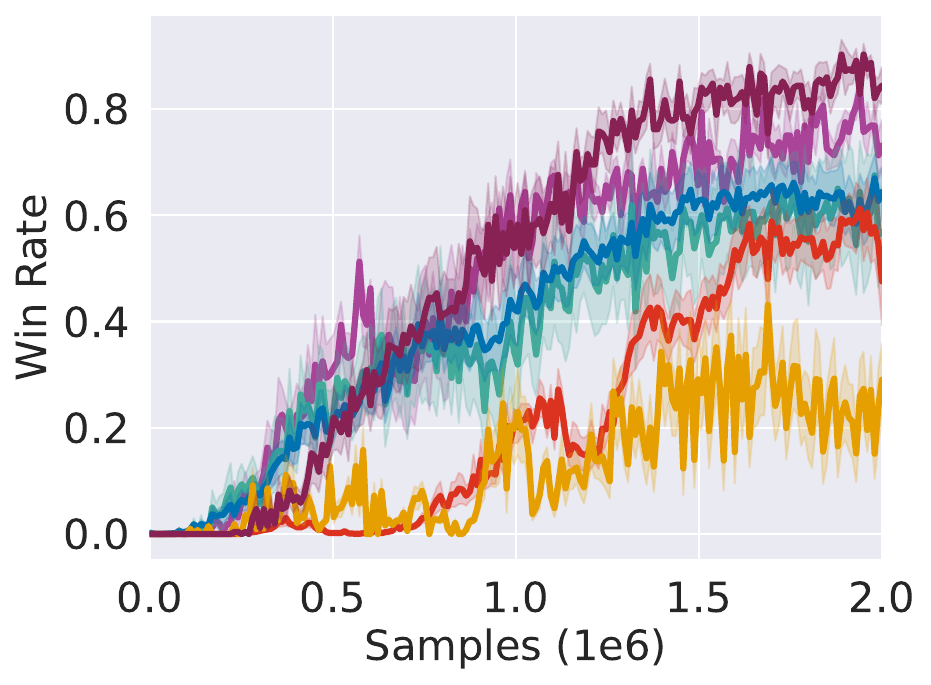}
        \caption{2c\_vs\_64zg}
        \label{fig:smac-2cv64zg}
    \end{subfigure}
    \begin{subfigure}[t]{0.24\textwidth}
        \centering
        \includegraphics[width=\textwidth]{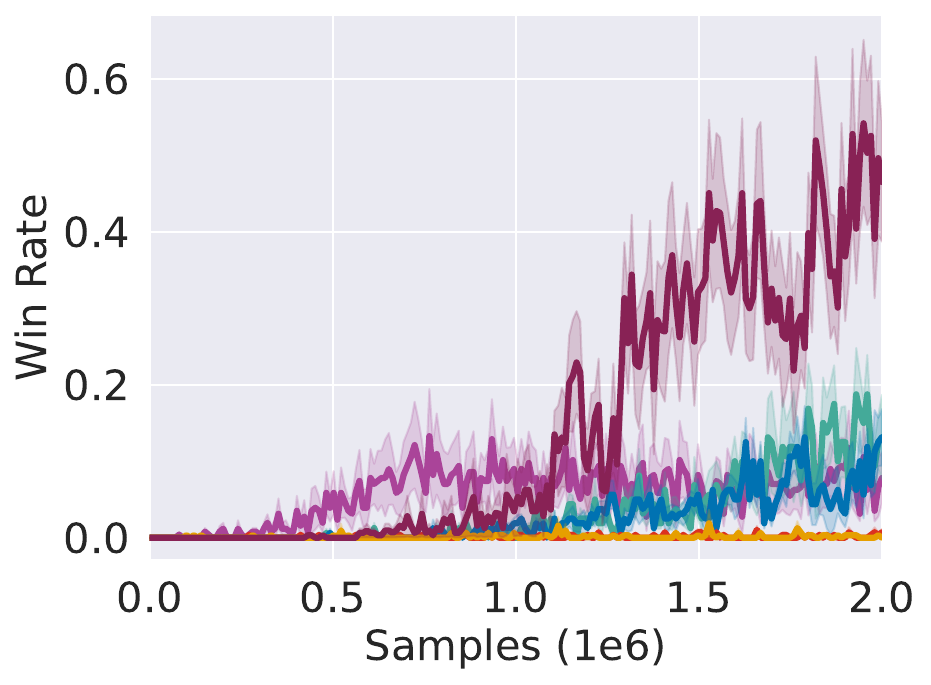}
        \caption{5m\_vs\_6m}
        \label{fig:smac-5mvs6m}
    \end{subfigure}
    \begin{subfigure}[t]{0.24\textwidth}
        \centering
        \includegraphics[width=\textwidth]{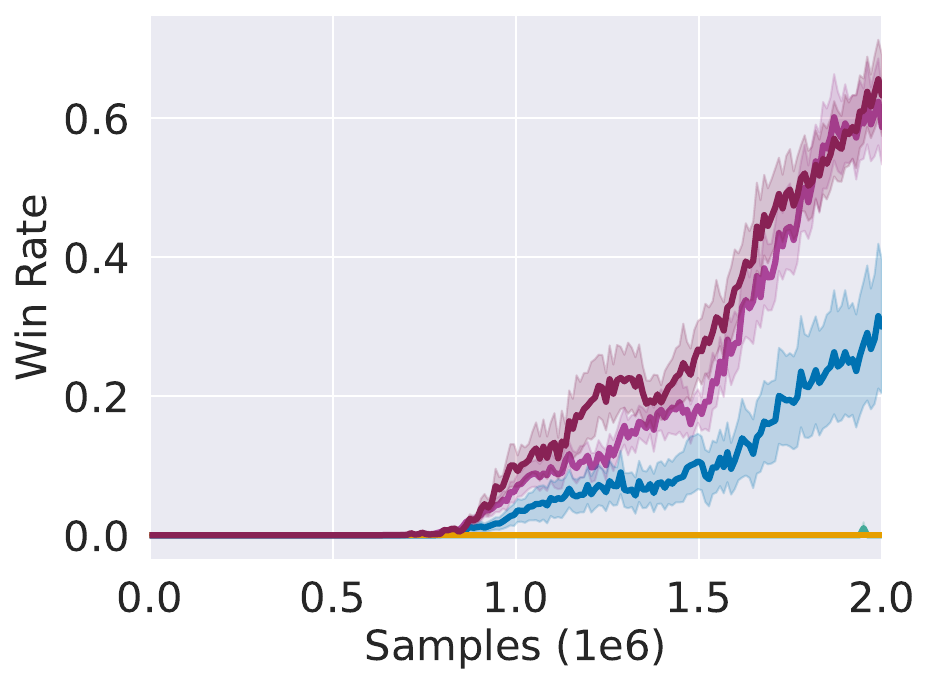}
        \caption{3s\_vs\_5z}
        \label{fig:smac-3svs5z}
    \end{subfigure}

    \vspace{0.2cm}
    
    \begin{subfigure}[t]{0.24\textwidth}
        \centering
    \includegraphics[width=\textwidth]{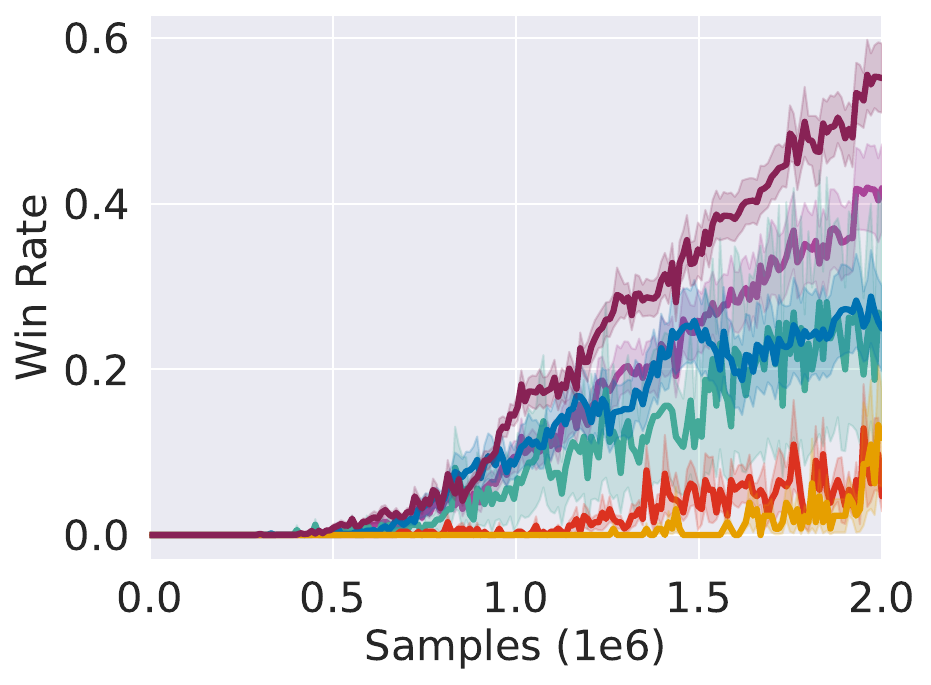}
        \caption{MMM2}
        \label{fig:smac-MMM2}
    \end{subfigure}
    \begin{subfigure}[t]{0.24\textwidth}
        \centering
        \includegraphics[width=\textwidth]{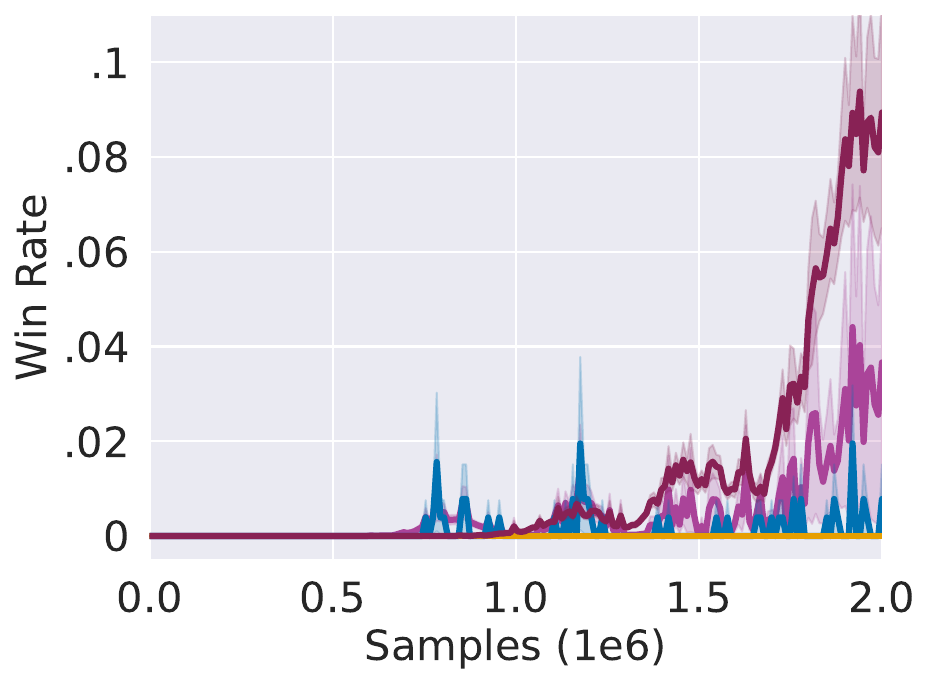}
        \caption{Corridor}
        \label{fig:smac-corridor}
    \end{subfigure}
    \begin{subfigure}[t]{0.24\textwidth}
        \centering
        \includegraphics[width=\textwidth]{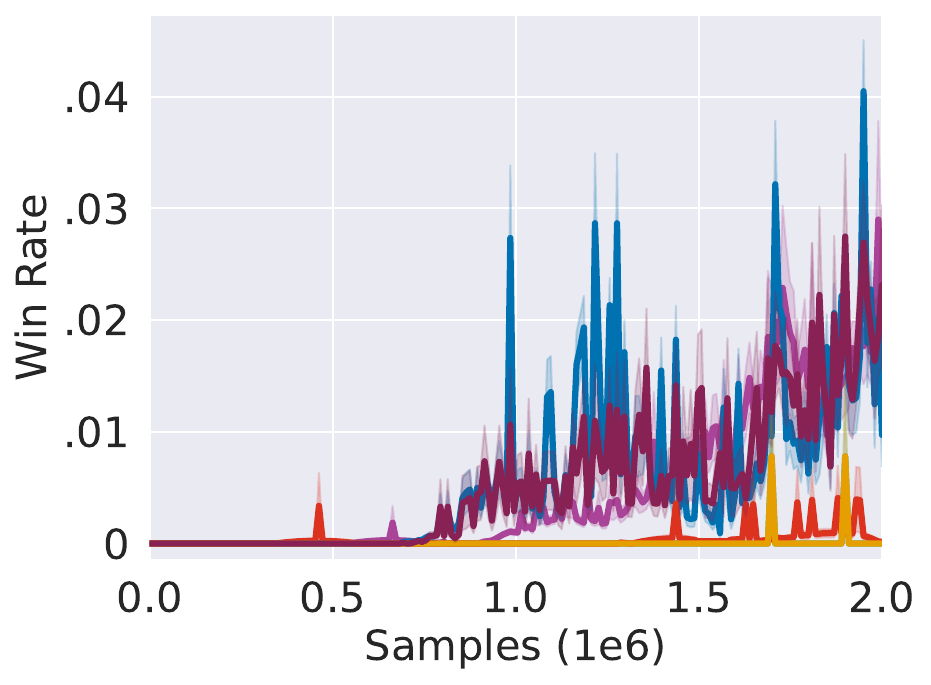}
        \caption{6h\_vs\_8z}
        \label{fig:smac-6hvs8z}
    \end{subfigure}
    \begin{subfigure}[t]{0.24\textwidth}
        \centering
        \includegraphics[width=\textwidth]{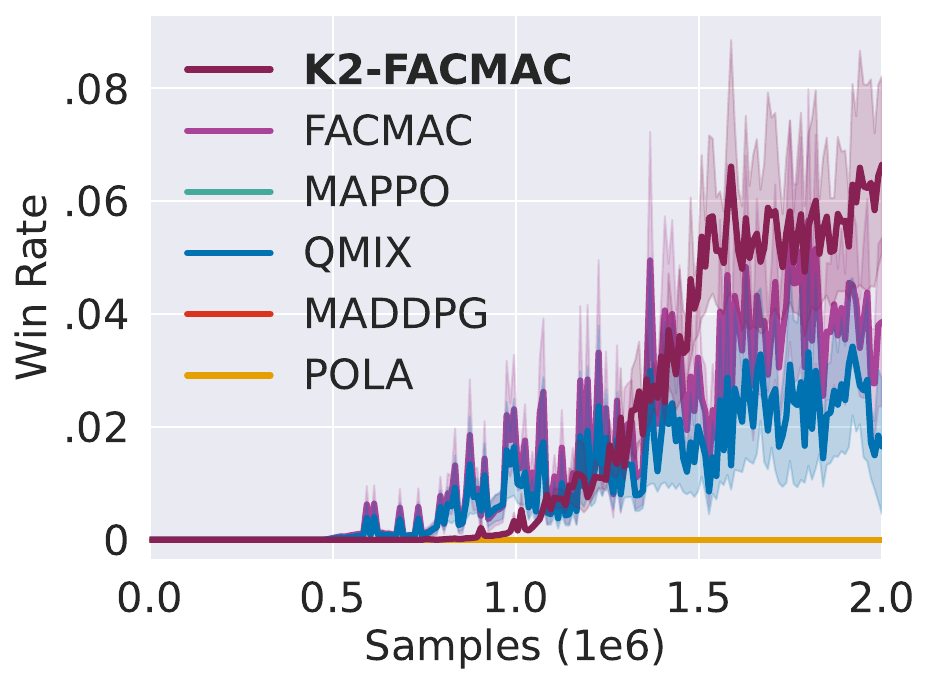}
        \caption{3s5z\_vs\_3s6z}
        \label{fig:smac-3s5zvs3s6z}
    \end{subfigure}
    \caption{Mean win rate and standard error of $K2$-FACMAC and baselines on SMAC maps across 10 seeds.}
    \label{fig:smac}
\end{figure*}

\begin{figure*}[t] 
    \centering
    \begin{subfigure}[t]{0.24\textwidth} 
        \centering
    \includegraphics[width=\textwidth]{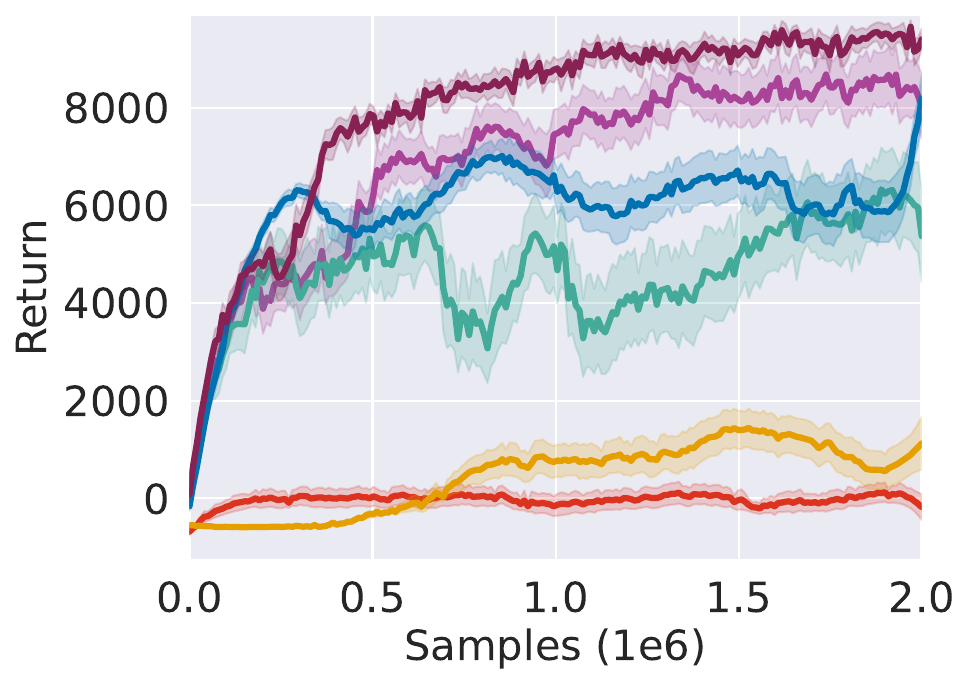}
        \caption{HalfCheetah-2x3}
        \label{fig:mm-halfcheetah2x3}
    \end{subfigure}
    \begin{subfigure}[t]{0.24\textwidth}
        \centering
        \includegraphics[width=\textwidth]{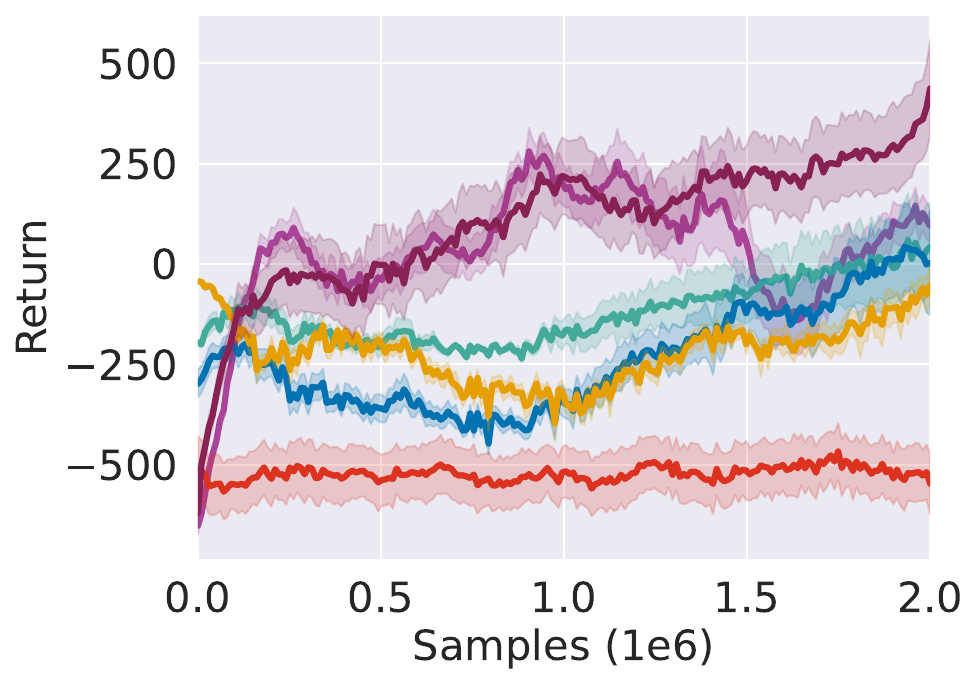}
        \caption{HalfCheetah-1p1}
        \label{fig:mm-halfcheetah1p1}
    \end{subfigure}
    \begin{subfigure}[t]{0.24\textwidth} 
        \centering
        \includegraphics[width=\textwidth]{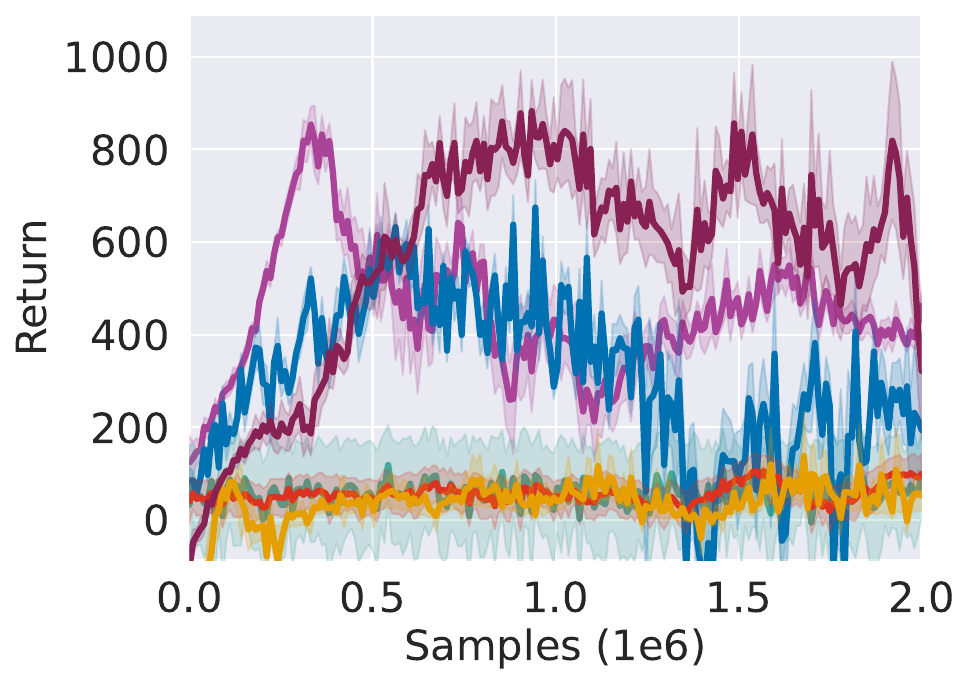}
        \caption{Ant-2x4}
        \label{fig:mm-ant}
    \end{subfigure}
    \begin{subfigure}[t]{0.24\textwidth} 
        \centering
        \includegraphics[width=\textwidth]{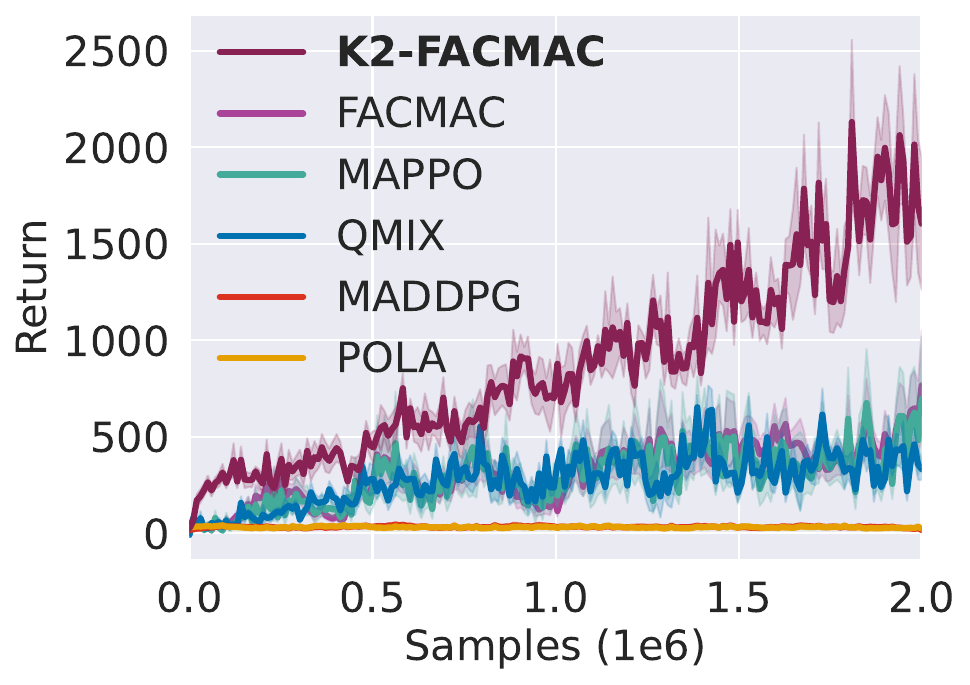}
        \caption{Walker-2x3}
        \label{fig:mm-walker}
    \end{subfigure}
    \caption{Mean performance and standard error of $K2$-FACMAC and baselines on four Multi-Agent MuJoCo environments across 10 seeds.}
    \label{fig:mm}
\end{figure*}

In this section, we demonstrate the effectiveness of KPG in deep MARL when applied to actor-critic algorithms. In the SMAX environment, we benchmark $K2$-MAPPO (MAPPO with $k{=}2$ levels of recursion) against several actor-critic and value-based algorithms. Due to the highly parallelized nature of SMAX~\citep{rutherford2023jaxmarl}, MAPPO performs well in terms of performance and wall-clock time, making it a good candidate to demonstrate the benefits of recursive reasoning.
In the SMAC and MAMuJoCo environments, we demonstrate the effectiveness of $K2$-FACMAC, as FACMAC achieves SOTA performance among off-policy multi-agent methods. In our experiments, we find that $k{=}2$ levels of policy recursion achieves most of the attainable performance benefits, with higher levels of recursion yielding small further improvements. 

\begin{figure*}[] 
    \centering
    \begin{subfigure}[t]{0.24\textwidth} 
        \centering
    \includegraphics[width=\textwidth]{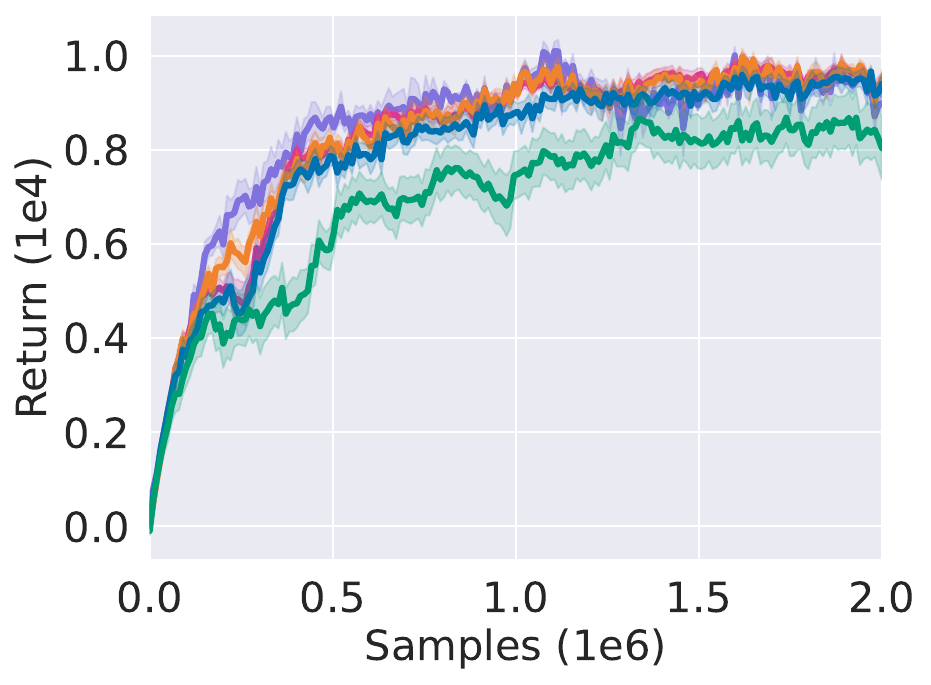}             
        \caption{\centering $K$-FACMAC\\(HalfCheetah-2x3)}
        \label{fig:ablate-k-facmac-mm}
    \end{subfigure}
    \begin{subfigure}[t]{0.24\textwidth}
        \centering
        \includegraphics[width=\textwidth]{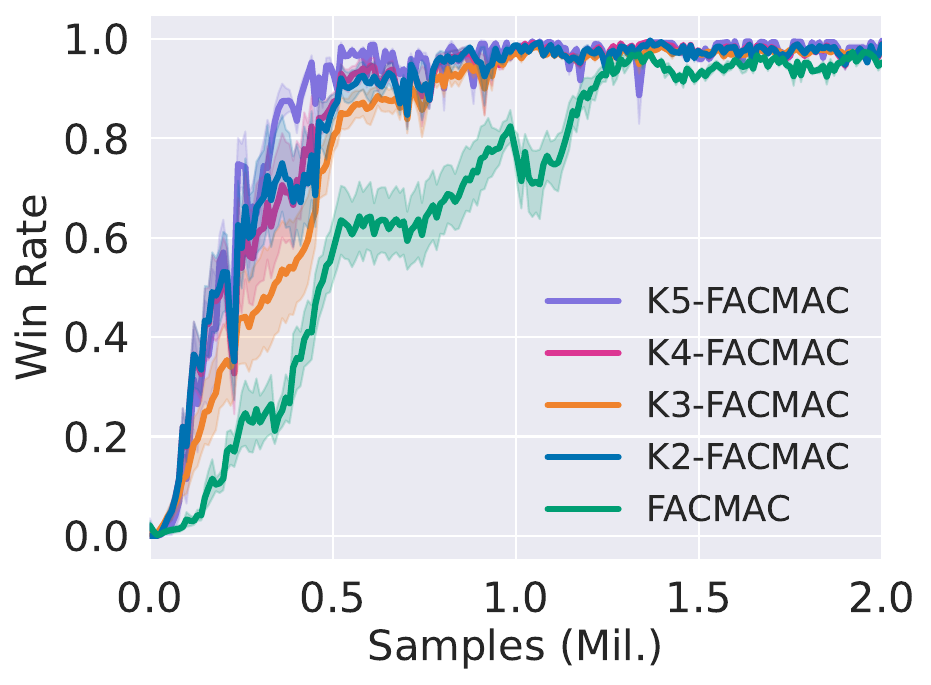}
        \caption{$K$-FACMAC (MMM)}
        \label{fig:ablate-k-facmac-smac}
    \end{subfigure}
    \begin{subfigure}[t]{0.24\textwidth} 
        \centering
        \includegraphics[width=\textwidth]{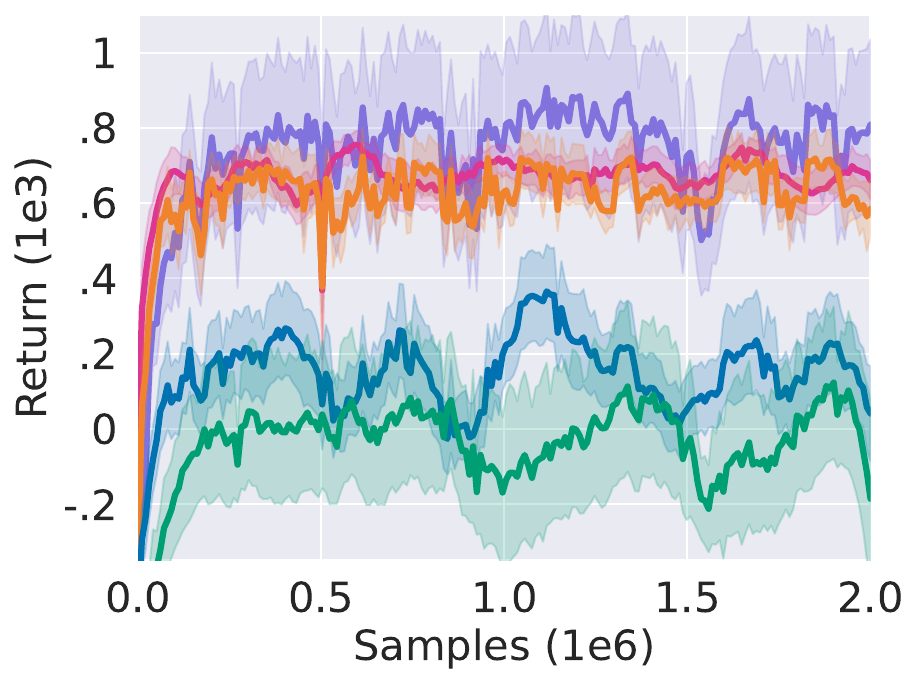}
        \caption{\centering $K$-MADDPG\\(HalfCheetah-2x3)}
        \label{fig:ablate-k-maddpg-mm}
    \end{subfigure}
    \begin{subfigure}[t]{0.24\textwidth} 
        \centering
        \includegraphics[width=\textwidth]{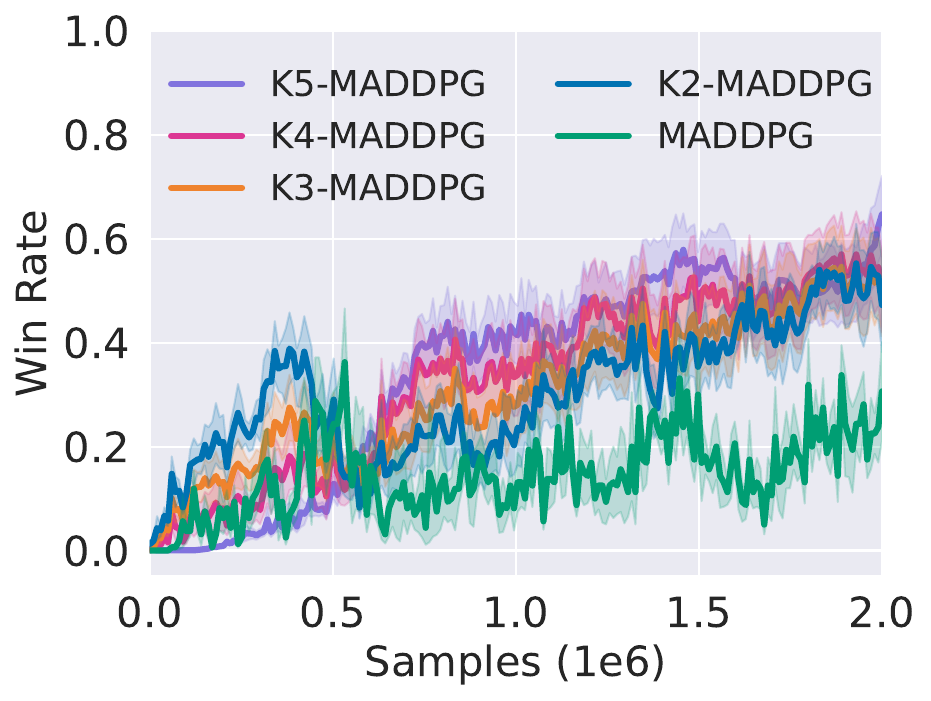}
        \caption{$K$-MADDPG (MMM)}
        \label{fig:ablate-k-maddpg-smac}
    \end{subfigure}
    \caption{Ablations for higher $K$-level FACMAC and MADDPG on MAMuJoCo (HalfCheetah-2x3) and SMAC (MMM).}
    \label{fig:ablate-k}
\end{figure*}

In addition to MAPPO and FACMAC, we benchmark against QMIX/COMIX due to their prevalence in MARL research and competitive performance, and POLA due to its use of higher-order opponent updates. We implement Outer-POLA with generalized advantage estimation as seen in~\citet{zhao2022proximal}. In SMAX, we also benchmark against VDN due to its relevance as a centralized value-based method and IPPO/IQL as examples of decentralized training. In SMAC and MAMuJoCo, we additionally benchmark against MADDPG with a centralized and monolithic critic. Note that our results may appear slightly different to previous works with SMAC/SMAX~\citep{rashid2020monotonic,rashid2020weighted} since previous works choose to report median instead of mean win rates, which reduces the impact of failing seeds, especially on certain difficult maps like Corridor.

\paragraph{K2-MAPPO outperforms related baselines in SMAX}
Figure~\ref{fig:smax} compares the test win rate of $K2$-MAPPO against related baselines on 11 SMAX maps (8 SMACv1-based and 3 SMACv2-based). Notably, $K2$-MAPPO performs equal or better than other baselines on 9 out of 11 maps, and is the only algorithm to solve 3s5z and 27m\_vs\_30m with 100\% accuracy. $K2$-MAPPO also maintains a consistent advantage over MAPPO, demonstrating the viability of recursive reasoning. The effects of recursive reasoning are seen in higher overall performance as well as faster convergence, as $K2$-MAPPO often reaches its maximum performance sooner than other methods. Despite utilizing opponent-shaping, POLA failed to match the performance of $K2$-MAPPO in every map except 3s5z\_vs\_3s6z. It is likely that the typical formulation of POLA, which performs well on 2-player reciprocity-based games, does not generalize well to more complex environments with many agents. 

\paragraph{K2-FACMAC outperforms related baselines in SMAC and MAMuJoCo} Figure ~\ref{fig:smac} compares the win rate of $K2$-FACMAC and related baselines across $8$ SMAC maps (Figure ~\ref{fig:smac}), with a focus on Hard and Super Hard maps. $K2$-FACMAC achieves higher or equal final success rates compared to baselines in every scenario. Using recursive reasoning, also achieves its maximum win rate sooner in maps which show equal final win rates (MMM, 3s\_vs\_5z, 6h\_vs\_8z). $K2$-FACMAC particularly stands out in Corridor (Figure~\ref{fig:smac-corridor}) and 3s5z\_vs\_3s6z (Figure~\ref{fig:smac-3s5zvs3s6z}), two notoriously difficult SMAC maps in which $K2$-FACMAC is the only algorithm to surpass a $5\%$ success rate. Note that MAPPO performs much worse on SMAC maps than SMAX maps due to SMAC being CPU-based and less parallelizable, resulting in far less sample availability (2e6 for each map rather than 1e7 in SMAX).

Figure~\ref{fig:mm} compares the performance of $K2$-FACMAC against the baselines on four selected MAMuJoCo environments. In all four environments, $K2$-FACMAC achieves superior performance at the end of training and tends to reach its peak performance earlier. In Ant 2x4 (Figure~\ref{fig:mm-ant}), learning a solution is difficult due to the asymmetric positioning of the agents which control opposing diagonal halves of a $4$-legged Ant agent. When observing the final learned policies, only $K2$-FACMAC, FACMAC, and COMIX demonstrate a positive improvement over the initial policy. $K2$-FACMAC and FACMAC both achieve the same level of maximum performance during training, with $K2$-FACMAC maintaining an advantage despite the unstable learning dynamics. $K2$-FACMAC also achieves the \textit{only} policy in Walker 2x3 (Figure~\ref{fig:mm-walker}) which adequately solves the task, obtaining SOTA performance on this very difficult locomotion benchmark. 

\paragraph{k=2 level KPG attains most of the available benefits} Figure~\ref{fig:ablate-k} compares the test performance of higher-level KPG applied to FACMAC and MADDPG upto $k{=}5$ in HalfCheetah-2x3 (MAMuJoCo) and MMM (SMAC). Note the performance increases over nominal MADDPG and FACMAC when KPG is applied, supporting the claim that our approach can reliably improve any centralized policy gradient algorithm. While the wall clock time increases fairly linearly as the number of policy recursions increases (Table~\ref{T:ablate-k} in Appendix~\ref{appendix:tables}), the additional benefits of increasing $K$ saturate quickly after $k=2$. Despite this, higher $k$-levels do exhibit monotonic performance increases and remain stable. Interestingly, Figure~\ref{fig:ablate-k-maddpg-mm} demonstrates a situation where higher-level recursions above $k=2$ are necessary for the algorithm to find a better policy mode.

\section{Conclusion and discussion}
We present $K$-Level Policy Gradients, an approach which harnesses the game theoretical paradigm of $k$-level thinking to improve the convergence of multi-agent policy gradient algorithms. We show that KPG in the limit (the Generalized Semi-Proximal Point Method) converges to a fixed point for N-player, general sum games. Furthermore, we show that our algorithm reaches an $\epsilon$-Nash equilibrium with a finite sequence of iterates under certain conditions. Our empirical results demonstrate KPG's ability to outperform existing MARL algorithms in the deep setting when applied to MAPPO, FACMAC, and MADDPG, enabling a further stepping stone to real-world applications of MARL. Future work in this area will consist of an analysis of KPG in the competitive setting and the applications of $k$-level thinking to other MARL approaches such as message passing. 

\textbf{Limitations.} The biggest practical limitation of KPG is its computational expense, since the number of backpropagation steps at each update scales linearly with $k$. In the future, we hope to reduce the computational expense of KPG using policy estimation methods such as opponent-policy modeling~\citep{he2016opponent} or parameter-based value estimation~\citep{faccio2020parameter}.

\clearpage
\section*{Acknowledgments}
This work was funded by the German Federal Ministry of Education and Research (BMBF) (Project: 01IS22078). This work was also funded by Hessian.ai through the project ’The Third Wave of Artificial Intelligence – 3AI’ by the Ministry for Science and Arts of the state of Hessen. 

Calculations for this research were conducted on the Lichtenberg high-performance cluster of TU Darmstadt, and the Intelligent Autonomous Systems (IAS) cluster
at TU Darmstadt. The authors also gratefully acknowledge the scientific support and HPC resources provided by the Erlangen National High Performance Computing Center (NHR@FAU) of the Friedrich-Alexander-Universität Erlangen-Nürnberg (FAU) under the NHR project b187cb. NHR funding is provided by federal and Bavarian state authorities. NHR@FAU hardware is partially funded by the German Research Foundation (DFG) -440719683.



\bibliography{kpg}

\begin{thebibliography}{52}
\providecommand{\natexlab}[1]{#1}
\providecommand{\url}[1]{\texttt{#1}}
\expandafter\ifx\csname urlstyle\endcsname\relax
  \providecommand{\doi}[1]{doi: #1}\else
  \providecommand{\doi}{doi: \begingroup \urlstyle{rm}\Url}\fi

\bibitem[Ismail et~al.(2018)Ismail, Sariff, and Hurtado]{ismail2018survey}
Zool~Hilmi Ismail, Nohaidda Sariff, and E~Gorrostieta Hurtado.
\newblock A survey and analysis of cooperative multi-agent robot systems: challenges and directions.
\newblock \emph{Applications of Mobile Robots}, 5:\penalty0 8--14, 2018.

\bibitem[Haydari and Y{\i}lmaz(2020)]{haydari2020deep}
Ammar Haydari and Yasin Y{\i}lmaz.
\newblock Deep reinforcement learning for intelligent transportation systems: A survey.
\newblock \emph{IEEE Transactions on Intelligent Transportation Systems}, 23\penalty0 (1):\penalty0 11--32, 2020.

\bibitem[Pi et~al.(2024)Pi, Zhang, Zhang, Huang, Rao, Ding, and Yang]{pi2024applications}
Yue Pi, Wang Zhang, Yong Zhang, Hairong Huang, Baoquan Rao, Yulong Ding, and Shuanghua Yang.
\newblock Applications of multi-agent deep reinforcement learning communication in network management: A survey.
\newblock \emph{arXiv preprint arXiv:2407.17030}, 2024.

\bibitem[Mnih et~al.(2015)Mnih, Kavukcuoglu, Silver, Rusu, Veness, Bellemare, Graves, Riedmiller, Fidjeland, Ostrovski, et~al.]{mnih2015human}
Volodymyr Mnih, Koray Kavukcuoglu, David Silver, Andrei~A Rusu, Joel Veness, Marc~G Bellemare, Alex Graves, Martin Riedmiller, Andreas~K Fidjeland, Georg Ostrovski, et~al.
\newblock Human-level control through deep reinforcement learning.
\newblock \emph{nature}, 518\penalty0 (7540):\penalty0 529--533, 2015.

\bibitem[Tang et~al.(2024)Tang, Abbatematteo, Hu, Chandra, Mart{\'\i}n-Mart{\'\i}n, and Stone]{tang2024deep}
Chen Tang, Ben Abbatematteo, Jiaheng Hu, Rohan Chandra, Roberto Mart{\'\i}n-Mart{\'\i}n, and Peter Stone.
\newblock Deep reinforcement learning for robotics: A survey of real-world successes.
\newblock \emph{Annual Review of Control, Robotics, and Autonomous Systems}, 8, 2024.

\bibitem[Li et~al.(2009)Li, Liao, and Carin]{Li2009Multi-task}
Hui Li, X.~Liao, and L.~Carin.
\newblock Multi-task reinforcement learning in partially observable stochastic environments.
\newblock \emph{J. Mach. Learn. Res.}, 10:\penalty0 1131--1186, 2009.
\newblock \doi{10.5555/1577069.1577109}.

\bibitem[Barfuss and Mann(2021)]{Barfuss2021Modeling}
W.~Barfuss and R.~Mann.
\newblock Modeling the effects of environmental and perceptual uncertainty using deterministic reinforcement learning dynamics with partial observability.
\newblock \emph{Physical review. E}, 105 3-1:\penalty0 034409, 2021.
\newblock \doi{10.1103/PhysRevE.105.034409}.

\bibitem[Robertson(1936)]{robertson1936general}
D~Robertson.
\newblock General theory of employment, interest and money.
\newblock \emph{QJ Econ}, 51:\penalty0 791--795, 1936.

\bibitem[Camerer et~al.(2004)Camerer, Ho, and Chong]{camerer2004cognitive}
Colin~F Camerer, Teck-Hua Ho, and Juin-Kuan Chong.
\newblock A cognitive hierarchy model of games.
\newblock \emph{The Quarterly Journal of Economics}, 119\penalty0 (3):\penalty0 861--898, 2004.

\bibitem[Gurney et~al.(2021)Gurney, Marsella, Ustun, and Pynadath]{gurney2021operationalizing}
Nikolos Gurney, Stacy Marsella, Volkan Ustun, and David~V Pynadath.
\newblock Operationalizing theories of theory of mind: a survey.
\newblock In \emph{AAAI Fall Symposium}, pages 3--20. Springer, 2021.

\bibitem[Schaafsma et~al.(2015)Schaafsma, Pfaff, Spunt, and Adolphs]{schaafsma2015deconstructing}
Sara~M Schaafsma, Donald~W Pfaff, Robert~P Spunt, and Ralph Adolphs.
\newblock Deconstructing and reconstructing theory of mind.
\newblock \emph{Trends in cognitive sciences}, 19\penalty0 (2):\penalty0 65--72, 2015.

\bibitem[Rutherford et~al.(2023)Rutherford, Ellis, Gallici, Cook, Lupu, Ingvarsson, Willi, Khan, de~Witt, Souly, et~al.]{rutherford2023jaxmarl}
Alexander Rutherford, Benjamin Ellis, Matteo Gallici, Jonathan Cook, Andrei Lupu, Gardar Ingvarsson, Timon Willi, Akbir Khan, Christian~Schroeder de~Witt, Alexandra Souly, et~al.
\newblock Jaxmarl: Multi-agent rl environments in jax.
\newblock \emph{arXiv preprint arXiv:2311.10090}, 2023.

\bibitem[Samvelyan et~al.(2019)Samvelyan, Rashid, De~Witt, Farquhar, Nardelli, Rudner, Hung, Torr, Foerster, and Whiteson]{samvelyan2019starcraft}
Mikayel Samvelyan, Tabish Rashid, Christian~Schroeder De~Witt, Gregory Farquhar, Nantas Nardelli, Tim~GJ Rudner, Chia-Man Hung, Philip~HS Torr, Jakob Foerster, and Shimon Whiteson.
\newblock The starcraft multi-agent challenge.
\newblock \emph{arXiv preprint arXiv:1902.04043}, 2019.

\bibitem[Peng et~al.(2021)Peng, Rashid, Schroeder~de Witt, Kamienny, Torr, B{\"o}hmer, and Whiteson]{peng2021facmac}
Bei Peng, Tabish Rashid, Christian Schroeder~de Witt, Pierre-Alexandre Kamienny, Philip Torr, Wendelin B{\"o}hmer, and Shimon Whiteson.
\newblock Facmac: Factored multi-agent centralised policy gradients.
\newblock \emph{Advances in Neural Information Processing Systems}, 34:\penalty0 12208--12221, 2021.

\bibitem[Yu et~al.(2022)Yu, Velu, Vinitsky, Gao, Wang, Bayen, and Wu]{yu2103surprising}
Chao Yu, Akash Velu, Eugene Vinitsky, Jiaxuan Gao, Yu~Wang, Alexandre Bayen, and Yi~Wu.
\newblock The surprising effectiveness of ppo in cooperative multi-agent games.
\newblock \emph{Advances in neural information processing systems}, 35:\penalty0 24611--24624, 2022.

\bibitem[Lowe et~al.(2017)Lowe, Wu, Tamar, Harb, Pieter~Abbeel, and Mordatch]{lowe2017multi}
Ryan Lowe, Yi~I Wu, Aviv Tamar, Jean Harb, OpenAI Pieter~Abbeel, and Igor Mordatch.
\newblock Multi-agent actor-critic for mixed cooperative-competitive environments.
\newblock \emph{Advances in neural information processing systems}, 30, 2017.

\bibitem[Kraemer and Banerjee(2016)]{kraemer2016multi}
Landon Kraemer and Bikramjit Banerjee.
\newblock Multi-agent reinforcement learning as a rehearsal for decentralized planning.
\newblock \emph{Neurocomputing}, 190:\penalty0 82--94, 2016.

\bibitem[Foerster et~al.(2018)Foerster, Farquhar, Afouras, Nardelli, and Whiteson]{foerster2018counterfactual}
Jakob Foerster, Gregory Farquhar, Triantafyllos Afouras, Nantas Nardelli, and Shimon Whiteson.
\newblock Counterfactual multi-agent policy gradients.
\newblock In \emph{Proceedings of the AAAI conference on artificial intelligence}, volume~32, 2018.

\bibitem[Du et~al.(2019)Du, Han, Fang, Liu, Dai, and Tao]{du2019liir}
Yali Du, Lei Han, Meng Fang, Ji~Liu, Tianhong Dai, and Dacheng Tao.
\newblock Liir: Learning individual intrinsic reward in multi-agent reinforcement learning.
\newblock \emph{Advances in Neural Information Processing Systems}, 32, 2019.

\bibitem[Sunehag et~al.(2017)Sunehag, Lever, Gruslys, Czarnecki, Zambaldi, Jaderberg, Lanctot, Sonnerat, Leibo, Tuyls, et~al.]{sunehag2017value}
Peter Sunehag, Guy Lever, Audrunas Gruslys, Wojciech~Marian Czarnecki, Vinicius Zambaldi, Max Jaderberg, Marc Lanctot, Nicolas Sonnerat, Joel~Z Leibo, Karl Tuyls, et~al.
\newblock Value-decomposition networks for cooperative multi-agent learning.
\newblock \emph{arXiv preprint arXiv:1706.05296}, 2017.

\bibitem[Son et~al.(2019)Son, Kim, Kang, Hostallero, and Yi]{son2019qtran}
Kyunghwan Son, Daewoo Kim, Wan~Ju Kang, David~Earl Hostallero, and Yung Yi.
\newblock Qtran: Learning to factorize with transformation for cooperative multi-agent reinforcement learning.
\newblock In \emph{International conference on machine learning}, pages 5887--5896. PMLR, 2019.

\bibitem[Rashid et~al.(2020{\natexlab{a}})Rashid, Farquhar, Peng, and Whiteson]{rashid2020weighted}
Tabish Rashid, Gregory Farquhar, Bei Peng, and Shimon Whiteson.
\newblock Weighted qmix: Expanding monotonic value function factorisation for deep multi-agent reinforcement learning.
\newblock \emph{Advances in neural information processing systems}, 33:\penalty0 10199--10210, 2020{\natexlab{a}}.

\bibitem[Rashid et~al.(2020{\natexlab{b}})Rashid, Samvelyan, De~Witt, Farquhar, Foerster, and Whiteson]{rashid2020monotonic}
Tabish Rashid, Mikayel Samvelyan, Christian~Schroeder De~Witt, Gregory Farquhar, Jakob Foerster, and Shimon Whiteson.
\newblock Monotonic value function factorisation for deep multi-agent reinforcement learning.
\newblock \emph{Journal of Machine Learning Research}, 21\penalty0 (178):\penalty0 1--51, 2020{\natexlab{b}}.

\bibitem[Wang et~al.(2020)Wang, Ren, Liu, Yu, and Zhang]{wang2020qplex}
Jianhao Wang, Zhizhou Ren, Terry Liu, Yang Yu, and Chongjie Zhang.
\newblock Qplex: Duplex dueling multi-agent q-learning.
\newblock \emph{arXiv preprint arXiv:2008.01062}, 2020.

\bibitem[Pan et~al.(2021)Pan, Rashid, Peng, Huang, and Whiteson]{pan2021softmax}
Ling Pan, Tabish Rashid, Bei Peng, Longbo Huang, and Shimon Whiteson.
\newblock Softmax with regularization: Better value estimation in multi-agent reinforcement learning.
\newblock \emph{arXiv preprint arXiv:2103.11883}, 2021.

\bibitem[Foerster et~al.(2017)Foerster, Chen, Al-Shedivat, Whiteson, Abbeel, and Mordatch]{foerster2017learning}
Jakob~N Foerster, Richard~Y Chen, Maruan Al-Shedivat, Shimon Whiteson, Pieter Abbeel, and Igor Mordatch.
\newblock Learning with opponent-learning awareness.
\newblock \emph{arXiv preprint arXiv:1709.04326}, 2017.

\bibitem[Willi et~al.(2022)Willi, Letcher, Treutlein, and Foerster]{willi2022cola}
Timon Willi, Alistair~Hp Letcher, Johannes Treutlein, and Jakob Foerster.
\newblock Cola: consistent learning with opponent-learning awareness.
\newblock In \emph{International Conference on Machine Learning}, pages 23804--23831. PMLR, 2022.

\bibitem[Zhao et~al.(2022)Zhao, Lu, Grosse, and Foerster]{zhao2022proximal}
Stephen Zhao, Chris Lu, Roger~B Grosse, and Jakob Foerster.
\newblock Proximal learning with opponent-learning awareness.
\newblock \emph{Advances in Neural Information Processing Systems}, 35:\penalty0 26324--26336, 2022.

\bibitem[Khan et~al.(2023)Khan, Willi, Kwan, Tacchetti, Lu, Grefenstette, Rockt{\"a}schel, and Foerster]{khan2023scaling}
Akbir Khan, Timon Willi, Newton Kwan, Andrea Tacchetti, Chris Lu, Edward Grefenstette, Tim Rockt{\"a}schel, and Jakob Foerster.
\newblock Scaling opponent shaping to high dimensional games.
\newblock \emph{arXiv preprint arXiv:2312.12568}, 2023.

\bibitem[Li et~al.(2016)Li, Oyler, Zhang, Yildiz, Girard, and Kolmanovsky]{li2016hierarchical}
Nan Li, Dave Oyler, Mengxuan Zhang, Yildiray Yildiz, Anouck Girard, and Ilya Kolmanovsky.
\newblock Hierarchical reasoning game theory based approach for evaluation and testing of autonomous vehicle control systems.
\newblock In \emph{2016 IEEE 55th Conference on Decision and Control (CDC)}, pages 727--733. IEEE, 2016.

\bibitem[Garz{\'o}n and Spalanzani(2019)]{garzon2019game}
Mario Garz{\'o}n and Anne Spalanzani.
\newblock Game theoretic decision making for autonomous vehicles’ merge manoeuvre in high traffic scenarios.
\newblock In \emph{2019 IEEE Intelligent Transportation Systems Conference (ITSC)}, pages 3448--3453. IEEE, 2019.

\bibitem[Bouton et~al.(2020)Bouton, Nakhaei, Isele, Fujimura, and Kochenderfer]{bouton2020reinforcement}
Maxime Bouton, Alireza Nakhaei, David Isele, Kikuo Fujimura, and Mykel~J Kochenderfer.
\newblock Reinforcement learning with iterative reasoning for merging in dense traffic.
\newblock In \emph{2020 IEEE 23rd International Conference on Intelligent Transportation Systems (ITSC)}, pages 1--6. IEEE, 2020.

\bibitem[Albaba and Yildiz(2021)]{albaba2021driver}
Berat~Mert Albaba and Yildiray Yildiz.
\newblock Driver modeling through deep reinforcement learning and behavioral game theory.
\newblock \emph{IEEE Transactions on Control Systems Technology}, 30\penalty0 (2):\penalty0 885--892, 2021.

\bibitem[Wang et~al.(2022)Wang, Zhang, and Peng]{wang2022comprehensive}
Xinpeng Wang, Songan Zhang, and Huei Peng.
\newblock Comprehensive safety evaluation of highly automated vehicles at the roundabout scenario.
\newblock \emph{IEEE Transactions on Intelligent Transportation Systems}, 23\penalty0 (11):\penalty0 20873--20888, 2022.

\bibitem[Karimi et~al.(2023)Karimi, Karimi, and Vahidi]{karimi2023level}
Shahab Karimi, Arash Karimi, and Ardalan Vahidi.
\newblock Level-$ k $ reasoning, deep reinforcement learning, and monte carlo decision process for fast and safe automated lane change and speed management.
\newblock \emph{IEEE Transactions on Intelligent Vehicles}, 8\penalty0 (6):\penalty0 3556--3571, 2023.

\bibitem[Dai et~al.(2023)Dai, Bae, and Isele]{dai2023game}
Siyu Dai, Sangjae Bae, and David Isele.
\newblock Game theoretic decision making by actively learning human intentions applied on autonomous driving.
\newblock \emph{arXiv preprint arXiv:2301.09178}, 2023.

\bibitem[Cui et~al.(2021)Cui, Hu, Pineda, and Foerster]{cui2021k}
Brandon Cui, Hengyuan Hu, Luis Pineda, and Jakob Foerster.
\newblock K-level reasoning for zero-shot coordination in hanabi.
\newblock \emph{Advances in Neural Information Processing Systems}, 34:\penalty0 8215--8228, 2021.

\bibitem[Bard et~al.(2020)Bard, Foerster, Chandar, Burch, Lanctot, Song, Parisotto, Dumoulin, Moitra, Hughes, et~al.]{bard2020hanabi}
Nolan Bard, Jakob~N Foerster, Sarath Chandar, Neil Burch, Marc Lanctot, H~Francis Song, Emilio Parisotto, Vincent Dumoulin, Subhodeep Moitra, Edward Hughes, et~al.
\newblock The hanabi challenge: A new frontier for ai research.
\newblock \emph{Artificial Intelligence}, 280:\penalty0 103216, 2020.

\bibitem[Liu and Pavel(2022)]{liu2022recursive}
Zichu Liu and Lacra Pavel.
\newblock Recursive reasoning in minimax games: A level $ k $ gradient play method.
\newblock \emph{Advances in Neural Information Processing Systems}, 35:\penalty0 16903--16917, 2022.

\bibitem[Goodfellow et~al.(2020)Goodfellow, Pouget-Abadie, Mirza, Xu, Warde-Farley, Ozair, Courville, and Bengio]{goodfellow2020generative}
Ian Goodfellow, Jean Pouget-Abadie, Mehdi Mirza, Bing Xu, David Warde-Farley, Sherjil Ozair, Aaron Courville, and Yoshua Bengio.
\newblock Generative adversarial networks.
\newblock \emph{Communications of the ACM}, 63\penalty0 (11):\penalty0 139--144, 2020.

\bibitem[Puterman(2014)]{puterman2014markov}
Martin~L Puterman.
\newblock \emph{Markov decision processes: discrete stochastic dynamic programming}.
\newblock John Wiley \& Sons, 2014.

\bibitem[Littman(1994)]{littman1994markov}
Michael~L Littman.
\newblock Markov games as a framework for multi-agent reinforcement learning.
\newblock In \emph{Machine learning proceedings 1994}, pages 157--163. Elsevier, 1994.

\bibitem[Lyu et~al.(2021)Lyu, Xiao, Daley, and Amato]{lyu2021contrasting}
Xueguang Lyu, Yuchen Xiao, Brett Daley, and Christopher Amato.
\newblock Contrasting centralized and decentralized critics in multi-agent reinforcement learning.
\newblock \emph{arXiv preprint arXiv:2102.04402}, 2021.

\bibitem[Kakade and Langford(2002)]{kakade2002approximately}
Sham Kakade and John Langford.
\newblock Approximately optimal approximate reinforcement learning.
\newblock In \emph{Proceedings of the nineteenth international conference on machine learning}, pages 267--274, 2002.

\bibitem[Fudenberg and Levine(1998)]{fudenberg1998theory}
Drew Fudenberg and David~K Levine.
\newblock \emph{The theory of learning in games}, volume~2.
\newblock MIT press, 1998.

\bibitem[Schulman et~al.(2017)Schulman, Wolski, Dhariwal, Radford, and Klimov]{schulman2017proximal}
John Schulman, Filip Wolski, Prafulla Dhariwal, Alec Radford, and Oleg Klimov.
\newblock Proximal policy optimization algorithms.
\newblock \emph{arXiv preprint arXiv:1707.06347}, 2017.

\bibitem[He et~al.(2016)He, Boyd-Graber, Kwok, and Daum{\'e}~III]{he2016opponent}
He~He, Jordan Boyd-Graber, Kevin Kwok, and Hal Daum{\'e}~III.
\newblock Opponent modeling in deep reinforcement learning.
\newblock In \emph{International conference on machine learning}, pages 1804--1813. PMLR, 2016.

\bibitem[Faccio et~al.(2020)Faccio, Kirsch, and Schmidhuber]{faccio2020parameter}
Francesco Faccio, Louis Kirsch, and J{\"u}rgen Schmidhuber.
\newblock Parameter-based value functions.
\newblock \emph{arXiv preprint arXiv:2006.09226}, 2020.

\bibitem[Zhao et~al.(2023)Zhao, Yang, Wang, and Lee]{zhao2023local}
Yulai Zhao, Zhuoran Yang, Zhaoran Wang, and Jason~D Lee.
\newblock Local optimization achieves global optimality in multi-agent reinforcement learning.
\newblock In \emph{International Conference on Machine Learning}, pages 42200--42226. PMLR, 2023.

\bibitem[Li et~al.(2022)Li, Xie, and Lu]{li2022difference}
Yueheng Li, Guangming Xie, and Zongqing Lu.
\newblock Difference advantage estimation for multi-agent policy gradients.
\newblock In \emph{International Conference on Machine Learning}, pages 13066--13085. PMLR, 2022.

\bibitem[Kingma(2014)]{kingma2014adam}
Diederik~P Kingma.
\newblock Adam: A method for stochastic optimization.
\newblock \emph{arXiv preprint arXiv:1412.6980}, 2014.

\bibitem[De~Boer et~al.(2005)De~Boer, Kroese, Mannor, and Rubinstein]{de2005tutorial}
Pieter-Tjerk De~Boer, Dirk~P Kroese, Shie Mannor, and Reuven~Y Rubinstein.
\newblock A tutorial on the cross-entropy method.
\newblock \emph{Annals of operations research}, 134:\penalty0 19--67, 2005.

\end{thebibliography}
\bibliographystyle{unsrtnat}


\newpage
\appendix

\section{Proofs}
\subsection{Proof of Theorem~\ref{th:cauchy}}

Recall Assumption~\ref{ass:lipschitz} and Algorithm~\ref{A:KPG}. We first analyze the pattern in successive updates of $\boldsymbol{\theta}$ as $k$ increases.

Consider level $K=1$ KPG:

\begin{equation}
\boldsymbol{\theta}_{t,i}^{k=1} = \boldsymbol{\theta}_{t,i} + \eta_{i}\nabla_{\boldsymbol{\theta}_{i}} J(\boldsymbol{\theta}_{t,i},\boldsymbol{\theta}_{t,-i}) \quad \forall i\in \mathcal{I}.
\end{equation}

The jump between $\boldsymbol{\theta}_t^{k=1}$ and $\boldsymbol{\theta}_{t}$ is

\begin{equation}
\|\boldsymbol{\theta}_{t,i}^{k=1} - \boldsymbol{\theta}_{t,i}\| = \eta_{i}\|\nabla_{\boldsymbol{\theta}_{i}}J_{i}(\boldsymbol{\theta}_{i}, \boldsymbol{\theta}_{-i})\| \quad \forall i\in \mathcal{I}.
\end{equation}

Thus, for all agents

\begin{equation}
\begin{split}
\|\boldsymbol{\theta}_{t}^{k=1} - \boldsymbol{\theta}_{t}\| 
&\leq \sum_{i=1}^{n}\|\boldsymbol{\theta}_{t,i}^{k=1} - \boldsymbol{\theta}_{t,i}\|
= \sum_{i=1}^{n} \eta_{i}\|\nabla_{\boldsymbol{\theta}_{i}}J_{i}(\boldsymbol{\theta}_{i}, \boldsymbol{\theta}_{-i})\| \\
& \leq \eta \sum_{i=1}^{n} \|\nabla_{\boldsymbol{\theta}_{i}}J_{i}(\boldsymbol{\theta}_{i}, \boldsymbol{\theta}_{-i})\|\\
& \leq \eta n \,\underset{i}{\max}\|\nabla_{\boldsymbol{\theta}_{i}}J_{i}(\boldsymbol{\theta}_{i}, \boldsymbol{\theta}_{-i})\| \\
& \leq \eta n \nabla_{\text{max}}.
\end{split}
\end{equation}

Now consider level $K=2$ KPG:

\begin{equation}
\boldsymbol{\theta}_{t,i}^{k=2} = \boldsymbol{\theta}_{t,i} + \eta_{i}\nabla_{\boldsymbol{\theta}_{i}} J(\boldsymbol{\theta}_{t,i},\boldsymbol{\theta}_{t,-i}^{k=1}) \quad \forall i\in \mathcal{I}.
\end{equation}

The jump between $\boldsymbol{\theta}_t^{k=2}$ and $\boldsymbol{\theta}_{t}^{k=1}$ is

\begin{equation}
\begin{split}
\|\boldsymbol{\theta}_{t,i}^{k=2} - \boldsymbol{\theta}_{t,i}^{k=1}\| & = \|\boldsymbol{\theta_{t,i}} + \eta_{i}\nabla_{\boldsymbol{\theta}_{i}}J_{i}(\boldsymbol{\theta}_{t,i}, \boldsymbol{\theta}_{t,-i}^{k=1}) - \boldsymbol{\theta_{t,i}} - \eta_{i}\nabla_{\boldsymbol{\theta}_{i}}J_{i}(\boldsymbol{\theta}_{t,i}, \boldsymbol{\theta}_{t,-i})\| \\
& =\eta_{i}\|\nabla_{\boldsymbol{\theta}_{i}}J_{i}(\boldsymbol{\theta}_{t,i}, \boldsymbol{\theta}_{t,-i}^{k=1}) - \nabla_{\boldsymbol{\theta}_{i}}J_{i}(\boldsymbol{\theta}_{t,i}, \boldsymbol{\theta}_{t,-i})\| \\
& \leq \eta_{i} L_{i} \|\boldsymbol{\theta}_{t,-i}^{k=1} - \boldsymbol{\theta}_{t,-i}\|  \quad \forall i\in \mathcal{I}. \
\end{split}
\end{equation}

Thus, for all agents

\begin{equation}
    \begin{split}
        \|\boldsymbol{\theta}_{t}^{k=2} - \boldsymbol{\theta}_{t}^{k=1}\| 
        & \leq \sum_{i=1}^{n}\|\boldsymbol{\theta}_{t,i}^{k=2} - \boldsymbol{\theta}_{t,i}^{k=1}\| \\
        & \leq \sum_{i=1}^{n} \eta_{i}L_{i} \|\boldsymbol{\theta}_{t,-i}^{k=1} - \boldsymbol{\theta}_{t,-i}\| \\
        &  \leq \sum_{i=1}^{n} \eta_{i}L_{i} \sum_{j\neq i}\|\boldsymbol{\theta}_{t,j}^{k=1} - \boldsymbol{\theta}_{t,j}\| \\
        & = \sum_{i=1}^{n} \eta_{i}L_{i} \sum_{j\neq i} \eta_{j}\|\nabla_{\boldsymbol{\theta}_{j}}J_{j}(\boldsymbol{\theta}_{t,j},\boldsymbol{\theta}_{t,-j}) \| \\
        & \leq \sum_{i=1}^{n} \eta_{i}L_{i} (n-1)\eta \nabla_{max} \\
        & \leq \eta^{2} L n (n-1)\nabla_{max}.
    \end{split}
\end{equation}

Now consider level $K=3$ KPG:

\begin{equation}
\boldsymbol{\theta}_{t,i}^{k=3} = \boldsymbol{\theta}_{t,i} + \eta_{i}\nabla_{\boldsymbol{\theta}_{i}} J(\boldsymbol{\theta}_{t,i},\boldsymbol{\theta}_{t,-i}^{k=2}) \quad \forall i\in \mathcal{I}.
\end{equation}

The jump between $\boldsymbol{\theta}_t^{k=3}$ and $\boldsymbol{\theta}_{t}^{k=2}$ is

\begin{equation}
    \begin{split}
      \|\boldsymbol{\theta}_{t,i}^{k=3} - \boldsymbol{\theta}_{t,i}^{k=2}\|  & =\eta_{i}\|\nabla_{\boldsymbol{\theta}_{i}}J_{i}(\boldsymbol{\theta}_{t,i}, \boldsymbol{\theta}_{t,-i}^{k=2}) - \nabla_{\boldsymbol{\theta}_{i}}J_{i}(\boldsymbol{\theta}_{t,i}, \boldsymbol{\theta}_{t,-i}^{k=1})\| \\
& \leq \eta_{i} L_{i} \|\boldsymbol{\theta}_{t,-i}^{k=2} - \boldsymbol{\theta}_{t,-i}^{k=1}\|  \quad \forall i\in \mathcal{I}. \
    \end{split}
\end{equation}

Thus, for all agents

\begin{equation}
    \begin{split}
        \|\boldsymbol{\theta}_{t}^{k=3} - \boldsymbol{\theta}_{t}^{k=2}\| 
        & \leq \sum_{i=1}^{n}\|\boldsymbol{\theta}_{t,i}^{k=3} - \boldsymbol{\theta}_{t,i}^{k=2}\| \\
        & \leq \sum_{i=1}^{n}\eta_{i}L_{i} \|\boldsymbol{\theta}_{t,i}^{k=2} - \boldsymbol{\theta}_{t,i}^{k=1}\| \\
        & \leq \sum_{i=1}^{n}\eta_{i}L_{i} \sum_{j\neq i}\|\boldsymbol{\theta}_{t,j}^{k=2} - \boldsymbol{\theta}_{t,j}^{k=1}\| \\
        & \leq \sum_{i=1}^{n}\eta_{i}L_{i} \sum_{j\neq i}\eta_{j}L_{j}\|\boldsymbol{\theta}_{t,-j}^{k=1} - \boldsymbol{\theta}_{t,-j}\| \\
        & \leq \sum_{i=1}^{n}\eta_{i}L_{i} \sum_{j\neq i}\eta_{j}L_{j}\sum_{l\neq j}\|\boldsymbol{\theta}_{t,m}^{k=1} - \boldsymbol{\theta}_{t,m}\| \\
        & = \sum_{i=1}^{n}\eta_{i}L_{i} \sum_{j\neq i}\eta_{j}L_{j}\sum_{l\neq j}\eta_{m}\|\nabla_{\boldsymbol{\theta}_{m}}J_{m}(\boldsymbol{\theta}_{t,m}, \boldsymbol{\theta}_{t,-m})\| \\
        & =  \sum_{i=1}^{n}\eta_{i}L_{i} \sum_{j\neq i}\eta_{j}L_{j}\sum_{l\neq j}\eta_{m}\|\nabla_{\boldsymbol{\theta}_{m}}J_{m}(\boldsymbol{\theta}_{t,m}, \boldsymbol{\theta}_{t,-m})\| \\
        & \leq \eta^{3}L^{2}n(n-1)^{2}\nabla_{max}.
    \end{split}
\end{equation}

We see by induction that any consecutive states during the $K-level$ procedure are bounded by

\begin{equation}
    \begin{split}
        \|\boldsymbol{\theta}_{t}^{k} - \boldsymbol{\theta}_{t}^{k-1}\| 
        & \leq \eta (\eta L)^{k-1}n(n-1)^{k-1}\nabla_{max}.
    \end{split}
\end{equation}

Let $\eta < \frac{1}{L(n-1)}$ such that the difference between between two $K$-steps is a contraction. Consider the difference $\|\boldsymbol{\theta}_{t}^{a} - \boldsymbol{\theta}_{t}^{b}\|$, where $a>b>0$:

\begin{equation}
    \begin{split}
        \|\boldsymbol{\theta}_{t}^{k=a} - \boldsymbol{\theta}_{t}^{k=b}\| 
        & = \|\sum_{j=b+1}^{a}\left(\boldsymbol{\theta}_{t}^{k=j} - \boldsymbol{\theta}_{t}^{k=j-1}\right)\| \\
        & \leq \sum_{j=b+1}^{a}\|\boldsymbol{\theta}_{t}^{k=j} - \boldsymbol{\theta}_{t}^{k=j-1}\|\\
        & \leq \sum_{j=b+1}^{a}\eta (\eta L)^{j-1}n(n-1)^{j-1}\nabla_{max} \\
        & \leq n\nabla_{max}  \left(\sum_{j=b+1}^{a}(n-1)^{j-1}L^{j-1}\right)\eta \left(\sum_{j=b+1}^{a} \eta^{j-1}\right) \\
        & \leq n\nabla_{max} \left(\sum_{j=b+1}^{a}(n-1)^{j-1}L^{j-1}\right)\eta \left(\sum_{j=b+1}^{a} \eta^{b-1}\right) \approx \mathcal{O}(\eta^{b}).
    \end{split}
\end{equation}

Thus, for any $\epsilon > 0$, we can solve for b such that $\eta (\eta L)^{k-1}n(n-1)^{k-1}\nabla_{max} < \epsilon$, or

\begin{equation}
    \exists N\in\mathbb{N} : \forall \epsilon>0, (a>b>N \implies\|\boldsymbol{\theta}^{a} - \boldsymbol{\theta}^{b}\| < \epsilon).
\end{equation}

Hence $\{\boldsymbol{\theta}_{t}^{k}\}_{k=0}^{\infty}$ is a Cauchy sequence. Since $\boldsymbol{\theta}_{t}$ lies in a complete subspace of $\mathbb{R}^{\sum_{i}d_{i}}$, the Cauchy sequence has a limit: $\underset{k\rightarrow\infty}{\lim}\boldsymbol{\theta}_{t}^{k} = \boldsymbol{\theta}_{t}^{\infty}$. \qed

\subsection{Proof of Theorem~\ref{th:GSPPM}}

Let us define $\boldsymbol{\hat{\theta}}_{t,i} = \boldsymbol{\theta}_{t,i} - \boldsymbol{\theta}^{*}_i$ and $\boldsymbol{\hat{\theta}}_{t} = [\boldsymbol{\hat{\theta}}_{t,1},...\boldsymbol{\hat{\theta}}_{t,n} ]^T $ for all $i \in \mathcal{I}$. It follows by linearizing the system about the stationary point $\boldsymbol{\theta}^{*}$, 

\begin{equation}
    \begin{split}
        \boldsymbol{\hat{\theta}}_{t+1,i} 
        & = \boldsymbol{\theta}_{t,i} + \eta_{i}\nabla_{\boldsymbol{\theta}_{i}}J_{i}(\boldsymbol{\theta}_{i}\boldsymbol{\theta}_{-i}) - \boldsymbol{\theta}^{*} \\
        & \approx \begin{pmatrix} I + \eta_{i}\nabla_{\boldsymbol{\theta}_{i},\boldsymbol{\theta}_{i}}^{2}J_{i}(\boldsymbol{\theta}_{i}^*,\boldsymbol{\theta}_{-i}^*), \eta_{i}\nabla_{\boldsymbol{\theta}_{i},\boldsymbol{\theta}_{-i}}^{2}J_{i}(\boldsymbol{\theta}_{i}^*,\boldsymbol{\theta}_{-i}^*)
        \end{pmatrix}
        \begin{pmatrix}
        \boldsymbol{\hat{\theta}}_{t,i}\\
        \boldsymbol{\hat{\theta}}_{t+1,-i}  
        \end{pmatrix}  \quad \textit{First order Taylor expansion} \\
        &= \boldsymbol{\hat{\theta}}_{t,i} + \eta_{i} \boldsymbol{A}_{i}\boldsymbol{\hat{\theta}}_{t,i} + \eta_{i}\boldsymbol{B}_{i}\boldsymbol{\hat{\theta}}_{t+1,-i}  \\
        \therefore  \boldsymbol{\hat{\theta}}_{t+1}  &= \boldsymbol{\hat{\theta}}_{t} + \boldsymbol{\eta} \boldsymbol{A} \boldsymbol{\hat{\theta}}_{t} + \boldsymbol{\eta} \boldsymbol{B} \boldsymbol{D} \boldsymbol{\hat{\theta}}_{t+1}. 
    \end{split}
\end{equation}

By analyzing the distance $r_{t}$ of the GSPPM iterates from the stationary point,

\begin{equation}
    \begin{split}
         r_{t+1}^2 &= \|\boldsymbol{\hat{\theta}}_{t+1}\|^{2} \\
         &= \boldsymbol{\hat{\theta}}_{t}^T(I + \boldsymbol{\eta} \boldsymbol{A})^T(I - \boldsymbol{\eta} \boldsymbol{B} \boldsymbol{D} )^{-T}(I - \boldsymbol{\eta} \boldsymbol{B} \boldsymbol{D} )^{-1}(I + \boldsymbol{\eta} \boldsymbol{A})\boldsymbol{\hat{\theta}}_{t}\\
         &\leq \frac{\underset{max}{\lambda}(I + \boldsymbol{\eta} \boldsymbol{A})^2}{\underset{min}{\lambda}(I-\boldsymbol{\eta} \boldsymbol{B} \boldsymbol{D})^2}r_{t}^2.
    \end{split}
\end{equation}

Thus, for any $\{\eta_{i}\} $ satisfying $\frac{\underset{max}{\lambda}(I + \boldsymbol{\eta} \boldsymbol{A})^2}{\underset{min}{\lambda}(I-\boldsymbol{\eta} \boldsymbol{B} \boldsymbol{D})^2} < 1$, GSPPM iterates converge asymptotically to the local Nash equilibrium. \qed

\subsection{Proof of Theorem~\ref{th:KPG_k_convergence}}
Let us define $\boldsymbol{\hat{\theta}}_{t,i}^{k} = \boldsymbol{\theta}_{t,i}^{k} - \boldsymbol{\theta}^{*}_i$ and $\boldsymbol{\hat{\theta}}_{t}^{k} = [\boldsymbol{\hat{\theta}}^{k}_{t,1},...\boldsymbol{\hat{\theta}}_{t,n}^{k} ]^T $ for all $i \in \mathcal{I}$. It follows by linearizing the system about the stationary point $\boldsymbol{\theta}^{*}$, 

\begin{equation}
    \begin{split}
        \boldsymbol{\hat{\theta}}_{t+1,i}^{k}
        & = \boldsymbol{\theta}_{t,i} + \eta_{i}\nabla_{\boldsymbol{\theta}_{i}}J_{i}(\boldsymbol{\theta}_{t,i}\boldsymbol{\theta}_{t,-i}^{k-1}) - \boldsymbol{\theta}^{*} \\
        & \approx \begin{pmatrix} I + \eta_{i}\nabla_{\boldsymbol{\theta}_{i},\boldsymbol{\theta}_{i}}^{2}J_{i}(\boldsymbol{\theta}_{i}^*,\boldsymbol{\theta}_{-i}^*), \eta_{i}\nabla_{\boldsymbol{\theta}_{i},\boldsymbol{\theta}_{-i}}^{2}J_{i}(\boldsymbol{\theta}_{i}^*,\boldsymbol{\theta}_{-i}^*)
        \end{pmatrix}
        \begin{pmatrix}
        \boldsymbol{\hat{\theta}}_{t,i}\\
        \boldsymbol{\hat{\theta}}_{t,-i}^{k-1}
        \end{pmatrix}  \quad \textit{First order Taylor expansion} \\
        &= \boldsymbol{\hat{\theta}}_{t,i} + \eta_{i} \boldsymbol{A}_{i}\boldsymbol{\hat{\theta}}_{t,i} + \eta_{i}\boldsymbol{B}_{i}\boldsymbol{\hat{\theta}}_{t,-i}^{k-1}  \\
        \therefore  \boldsymbol{\hat{\theta}}_{t}^{k}  &= \boldsymbol{\hat{\theta}}_{t} + \boldsymbol{\eta} \boldsymbol{A} \boldsymbol{\hat{\theta}}_{t} + \boldsymbol{\eta} \boldsymbol{B} \boldsymbol{D} \boldsymbol{\hat{\theta}}_{t}^{k-1}.
    \end{split}
\end{equation}

By analyzing the distance $r_{t}^{k}$ of the KPG iterates from the stationary point,

\begin{equation}
    \begin{split}
         (r_{t}^{k})^2 &= \|\boldsymbol{\hat{\theta}}_{t}^{k}\|^{2} \\
         &= \boldsymbol{\hat{\theta}}_{t}^T(I+\boldsymbol{\eta} \boldsymbol{A})^T(I+\boldsymbol{\eta} \boldsymbol{A})\boldsymbol{\hat{\theta}}_{t} + \boldsymbol{\hat{\theta}}_{t}^T(I+\boldsymbol{\eta} \boldsymbol{A})^T\boldsymbol{\eta} \boldsymbol{B} \boldsymbol{D} \boldsymbol{\hat{\theta}}_{t}^{k-1} \\
         & + (\boldsymbol{\hat{\theta}}_{t}^{k-1})^T\boldsymbol{D}^T\boldsymbol{B}^T\boldsymbol{\eta}^T(I+\boldsymbol{\eta} \boldsymbol{A})\boldsymbol{\hat{\theta}}_{t} + (\boldsymbol{\hat{\theta}}_{t}^{k-1})^T\boldsymbol{D}^T\boldsymbol{B}^T\boldsymbol{\eta}^T\boldsymbol{\eta} \boldsymbol{B}\boldsymbol{D}\boldsymbol{\hat{\theta}}_{t}^{k-1}\\
         & \leq \left(\underset{max}{\lambda}(I + \boldsymbol{\eta} \boldsymbol{A})^2 + 2\underset{max}{\lambda}(I+\boldsymbol{\eta} \boldsymbol{A})\underset{max}{\lambda}(\boldsymbol{\eta} \boldsymbol{B}\boldsymbol{D}) \right)(r_{t}^{0})^2 + \\ & 2\underset{max}{\lambda}(I+\boldsymbol{\eta} \boldsymbol{A})\underset{max}{\lambda}(\boldsymbol{\eta} \boldsymbol{B}\boldsymbol{D})\left(r_{t}^{0}\nabla_{max} \right) + \underset{max}{\lambda}(\boldsymbol{\eta} \boldsymbol{B}\boldsymbol{D})^2 (r_{t}^{k-1})^2.
    \end{split}
\end{equation}

defining the bound of the finite-k iterates to the stationary point $\boldsymbol{\hat{\theta}}^{*}$.

Hence, for any $\{\eta_{i}\}$ satisfying $\underset{max}{\lambda}(\boldsymbol{\eta} \boldsymbol{B}\boldsymbol{D})^2 < 1$, KPG iterates converge asymptotically to the local Nash Equilibrium. \qed

\section{Details of the illustrative example}
\label{appendix:illustrative_example}
We report the full details of the toy problem introduced in Sec.~\ref{sec:illustrative_example}, which we here refer to as the \emph{Meet-up} problem.

\paragraph{Environment properties.} The problem is designed as a simple 2-player continuous cooperative game in  a 2D space. The state of the game \( s=(s_1, s_2) \in \mathbb{R}^4\) encodes the location of the two players, with \( s_i \in \mathbb{R}^2 \). For the sake of simplicity, agents can only move by a fixed distance step of 1 around their current position, towards a chosen direction.
The initial state of the two agents is deterministic and fixed to \( \iota = (\iota_1=(0,0), \iota_2=(3,2)) \). We assume undiscounted returns (\(\gamma=1\)) and terminate an episode when the agents effectively meet each other as a result of their actions.

\paragraph{Policy parameterization.} Although one-dimensional continuous actions are trivially tractable for one-step games, sequential decision making problems demand finding policies that respond optimally for any possible configuration \( s \) of the game. Here, we reduce the complexity of the problem by conveniently parameterizing each agent as single-parameter policies. In particular, we define an agent action as a 1-DoF unit vector \( a_i \in \mathbb{R}^2 \), and parameterize the deterministic policy of agent \( i \) with \( \theta_i \in \mathbb{R} \), as

\begin{equation}
    \pi _{i}(s) = \begin{cases}
(\cos\theta _{i} ,\sin\theta _{i})^\top & \mathrm{if,} \ s=\iota\\
\pi_i ^{*}( s) & \mathrm{if,} \ s\neq \iota
\end{cases}
\end{equation}

where, \(
    \pi_i^*(s) = \frac{s_{-i} - s_{i}}{\lVert s_{-i} - s_{i} \lVert} 
\) is the optimal policy that goes straight towards the other agent. 
In other words, we assume that both agents will act optimally after taking the first action, and we only parametrize the agents decisions at the starting state.
This design choice allows to easily study the joint policy space directly, as well as computing the closed-form solution of the return \( J(\cdot) \) (see below).

\paragraph{Solving the Meet-up problem with KPG.} We design the reward function of the Meet-up problem to reward each agent for getting closer to the other agent after the effect of both actions. We achieve this by computing the cosine similarity between the agent's action \( a_i \) and the actual direction that would have led closest to the other agent:
\begin{equation}
    R_i(s,a,s') = a_i \cdot \pi_i^*(s_i, s'_{-i}) - 1= a_i \cdot \frac{s'_{-i} - s_i}{\lVert s'_{-i} - s_i\lVert} - 1.
\end{equation}
Here, we denote the joint action as \( a = (a_1, a_2) \), and the next state as \( s' =  (s_1+a_1, s_2+a_2) \).
Note that a $-1$ offset is added so that both the reward signal and the return \( J_i(\theta_i, \theta_{-i}) \) of each agent is always \( \leq 0 \). In turn, this makes the computation of the optimal value function \( V^*(s) \) of this game trivial: the strategy of moving towards each other in a straight line leads to returns of $0$ from any state \( s \); since this is the maximum return, this joint policy must also be optimal, and \( V^*(s) = 0 \ \forall s\) is the unique optimal value function.
We now derive the analytical form of \( J_i(\theta_i, \theta_{-i})\), as needed by KPG to compute the recursive gradient updates. Given that both agents are assumed to act optimally in any state besides the starting state, we can conveniently write the return as
\begin{equation}
    \begin{split}
        J_i(\theta_i, \theta_{-i}) &=  R_i(\iota, a, s') + V_i(s')\\
        &= R_i(\iota, a, s') + V^*(s') = \\
        &= R_i(\iota, a, s')
    \end{split}
\end{equation}
where \( s' \) is the resulting state after the players' first actions \( a_1 = (\cos\theta_1, \sin\theta_1) \) and \( a_2 = (\cos\theta_2, \sin\theta_2)\).
Following this, we may therefore compute the gradient of the return for any pair of agent policies \( \theta_1, \theta_2 \) in closed form:

\begin{equation}
\label{eq:closed_form_return}
    \begin{split}
        \nabla_{\theta_i} J_i(\theta_i, \theta_{-i}) &= \nabla_{\theta_i} R_i(\iota, a, s') \\
        &= \nabla_{\theta_i} \Bigl( \ a_i \cdot \pi_i^*(\iota_i, s'_{-i}) \ \Bigl) \\
        &= \nabla_{\theta_i} \Bigl( a_i\Bigl) \cdot \pi_i^*(\iota_i, s'_{-i}) \\
        &= \nabla_{\theta_i} \begin{pmatrix}
            \cos\theta_i \\
            \sin\theta_i
            \end{pmatrix} \cdot \pi_i^*(\iota_i, s'_{-i}) \\
        &= \begin{pmatrix}
            -\sin\theta_i \\
            \cos\theta_i
            \end{pmatrix} \cdot \pi_i^*(\iota_i, s'_{-i})
    \end{split}
\end{equation}
In conclusion, Eq.~\ref{eq:closed_form_return} allows us to compute the recursive $K$-level reasoning steps with the true analytical gradient of the return, as shown in Algorithm~\ref{A:KPG} of the main manuscript.

\section{Practical Implementation of the $K$-Level Policy Gradient}
\textbf{The performance difference lemma  (PDL)}~\citep{kakade2002approximately} \textbf{can be extended to the $k$-level multi-agent setting}~\citep{zhao2023local,li2022difference}.
\begin{lemma}
    
It can be shown that for agent $i$,

\begin{align}
    J_{i}(\pi_{i},\boldsymbol{\pi}^{(0)}_{-i}) - J_{i}(\pi^{(0)}_{i},\boldsymbol{\pi}^{(0)}_{-i}) 
    & = \frac{1}{1-\gamma}\mathbb{E}_{(s,\boldsymbol{a})\sim d^{\pi_i,\boldsymbol{\pi}^{(0)}_{-i}},\pi_{i},\boldsymbol{\pi}^{(0)}_{-i}}\left[ A_{i}^{\boldsymbol{\pi}^{(0)}}(s,\boldsymbol{a})\right] \\
    & = \frac{1}{1-\gamma}\mathbb{E}_{(s,\boldsymbol{a})\sim d^{\pi_i,\boldsymbol{\pi}^{(0)}_{-i}},\pi^{(0)}_{i},\boldsymbol{\pi}^{(0)}_{-i}}\left[ \frac{\pi_{i}}{\pi^{(0)}_{i}}A^{\boldsymbol{\pi}^{(0)}}(s,\boldsymbol{a})\right],
\end{align}

where $\pi_{i}$ is the policy of agent $i$, $\boldsymbol{\pi}^{(0)}_{-i}$ is the $0$-level joint policy of the other agents, $J_{i}(.)$ is agent $i$'s objective function, $\boldsymbol{a}$ is the joint action, $A^{(0)} _{i}(s,\boldsymbol{a})$ is agent $i$'s advantage function, and $d^{(.)}$ is the discounted state distribution. This is the version of the PDL used by MAPPO; since it does not consider the update of the other agents, they are treated like part of the environment.

Now consider the PDL when updating against $k$-level agents:

\begin{align}
    J_{i}(\pi_{i},\boldsymbol{\pi}^{(k)}_{-i}) - J_{i}(\pi^{(0)}_{i},\boldsymbol{\pi}^{(0)}_{-i}) 
    & = \frac{1}{1-\gamma}\mathbb{E}_{(s,\boldsymbol{a})\sim d^{\pi_i,\boldsymbol{\pi}^{(k)}_{-i}},\pi_{i},\boldsymbol{\pi}^{(k)}_{-i}}\left[ A_{i}^{\boldsymbol{\pi}^{(0)}}(s,\boldsymbol{a})\right] \\
    & = \frac{1}{1-\gamma}\mathbb{E}_{(s,\boldsymbol{a})\sim d^{\pi_i,\boldsymbol{\pi}^{(0)}_{-i}},\pi^{(0)}_{i},\boldsymbol{\pi}^{(0)}_{-i}}\left[ \frac{\pi_{i}}{\pi^{(0)}_{i}} \frac{\boldsymbol{\pi}^{(k)}_{-i}}{\boldsymbol{\pi}^{(0)}_{-i}} A^{\boldsymbol{\pi}^{(0)}}(s,\boldsymbol{a})\right],
\end{align}

which includes a correction ratio accounting for the change in action distribution of the $k$-level agents. Note that both cases feature a mismatch in the discounted state distribution $d$ of the PDL and the sample distribution ($d^{\pi_i,\boldsymbol{\pi}_{-i}}$); this mismatch is ignored in practice.
\end{lemma}

\textbf{The deterministic policy gradient can be estimated for $k$-level policies.} Centralized deterministic policy gradient algorithms utilize an action-value function trained using off-policy transitions. In our implementation of $K$-MADDPG and $K$-FACMAC, we use an off-policy critic $Q_{i}^{\boldsymbol{\pi}^{(0)}}(s,.,.)$, which estimates the action-value function for agent $i$ under the $k=0$ policies of all the agents. Since the critic is trained using off-policy data, it is reasonably well-estimated for higher $k$-level updates which are obtained by estimating $Q_{i}^{\boldsymbol{\pi}^{(0)}}(s,a_{i},\boldsymbol{a}_{-i}^{(k)})$ for each agent $i$, since the joint $k$-level actions $\boldsymbol{a}_{-i}^{(k)}$ are only obtained by policies one gradient step away from $\boldsymbol{\pi}_{-i}^{(0)}$. This is similar to~\citet{peng2021facmac}, where the centralized critic is trained using off-policy actions and used to estimate the action-value function for the online actions of all agents to prevent relative overgeneralization.




\section{Environment Details}
All SMAX maps were configured according to the default settings seen in~\citet{rutherford2023jaxmarl}. All MAMuJoCo environments and agents are configured according to the default configurations used in~\citet{peng2021facmac} where they were introduced. Each agent observes the positions of its own body parts, receives a common team reward that depends on the task, and controls only its joints. The exact configurations and rewards can be seen at \url{https://robotics.farama.org/envs/MaMuJoCo/}. All experiments using SMAC mostly used the default team configurations, rewards, and observations as the SMAC benchmark~\citep{samvelyan2019starcraft}. The state space was modified slightly by including the last actions of each agent (using the inbuilt feature in the StarCraft II environment) as this was found to stabilize learning for all algorithms.

\section{Experimental Details}
\subsection{SMAX}
Each baseline is run with the settings seen in~\citet{rutherford2023jaxmarl}. Each baseline uses $1e7$ total training steps and is trained against the `HeuristicEnemySMAX' AI, updated every 128 steps, and uses a $\gamma$ of $0.99$. \textbf{Off-policy algorithms} (QMIX, VDN, IQL) are trained with $16$ parallel environments with a buffer size of $5000$ and a batch size of $32$. Each uses Adam optimizers with a learning rate of $5e-5$, and performs $\epsilon$-greedy exploration during training time with and $\epsilon$ that decays from $1$ to $0.05$ over the first $10\%$ of total steps (learning is also paused until the $\epsilon$ decay is concluded). Neural networks use a hidden size of $512$ and relu activations, hard target updates every $10$ updates, $8$ update epochs, and a reward scale of $10$ (the reward scale of the original SMAC environments). The maximum gradient norm is constrained to be $10$. In QMIX, the mixer embedding dimension is $64$, the mixer hypernet hidden dimension is $256$, and the initial scale of the kernel weights of the mixer weights is set to $0.001$. Baselines are evaluated every $5\%$ of total steps for $128$ steps across $512$ environments. \textbf{On-policy algorithms} (K-MAPPO, MAPPO, IPPO, POLA) are trained with $64$ parallel environments. Each uses Adam optimizers with a learning rate of $4e-3$ which is annealed to $0$ over the entire course of training. Neural networks use a hidden size of $128$ and relu activations, $2$ minibatch updates, and $2$ update epochs. The maximum gradient norm is constrained to be $0.5$. The value of $\lambda$ for the GAE is set to $0.95$, the value of $\epsilon$ for surrogate clipping is $0.2$, the value loss coefficient is $0.5$, and no entropy bonus is provided.

\subsection{Multi-Agent MuJoCo and SMAC}
For each algorithm, we evaluate the performance in the following manner: we pause training after 10,000 steps and run a fixed number of independent test episodes (10 for MAMuJoCo and 32 for SMAC). During these test episodes, each agent acts greedily in a decentralized fashion (DDPG agents don't use action noise, DQN agents don't select random actions, etc.). The mean performance of the agents is reported in MAMuJoCo (the performance for each agent is identical since they share a common objective) and the mean success rate is reported for SMAC. Note that we chose to report the mean success rate rather than the median, as certain SMAC maps (especially Super Hard maps) commonly result in very success rates. Using the mean success rate better reflects the difficulty of these maps, as the median success rate can sometimes skew results to look more positive than they should. We set $\gamma=0.99$ for all experiments.

Since our results are primarily obtained by applying KPG to FACMAC, we mostly kept the algorithmic implementation standards used in FACMAC for reproducibility. We use parameter sharing for all actor and critic networks to speed up learning. All actor, critic, mixer, and Q-networks have target networks.

In torch environments, we find that RMSProp provides the most stable updates, since the momentum effects of Adam appear to mitigate the benefits of KPG by carrying over a running average of gradients from previous timesteps (which does not help when KPG's benefit is a better gradient for the \textit{current} timestep.) As a practical detail, note also that at the end of each non-final intermediate gradient step taken by KPG, the statistics of the optimizer must be reset to what they were at the beginning of that timestep, as the carryover effects of any optimizer from the previous timestep must apply to each $K$-level gradient independently. 

To clarify our implementation of $K$-FACMAC, we provide Algorithm~\ref{A:KFACMAC} below, which provides further details on $K$-level action sampling, intermediate fitting using optimizers, and resets of the actors and optimizers after each $K$-level step \textit{except} the final one.

\textbf{MAMuJoCo} The architecture of all deep Q-networks is an MLP with 2 hidden layers with 400 and 300 units respectively. In all actor-critic methods, the architecture of the shared actor and critic networks is an MLP with 2 hidden layers with 400 and 300 hidden units respectively. All hidden layers for all networks use ReLU activations. All critic networks provide raw outputs while actor networks have a tanh activation at the output. Actor networks and DQNs receive the local observations of that agent as an input, appended with a one-hot vector due to the parameter sharing. All centralized critics and mixing networks are conditioned on the global state provided by the environment. 

Each episode has a maximum length of 1000 steps. The total training time for each algorithm is set to 2 million steps. To improve initial exploration, each agent takes 10,000 random steps at the beginning of each run. During training we apply uncorrelated, mean-zero noise with a standard deviation of 0.1 to further encourage exploration. Each agent has a replay buffer with a maximum size of 1 million and trains with a batch size of 100 after every new sample. Target networks are updated using Polyak averaging with $\tau=0.001$. All neural networks are trained using the Adam optimizer~\citep{kingma2014adam} (except for KPG which uses RMSProp) with a learning rate of 0.001. 

\textbf{SMAC} The architecture of all shared deep Q-networks is a DRQN with a recurrent layer comprised of a GRU with a 64-dimensional hidden state, with fully connected layers on either side. In all actor-critic methods, the architecture of all shared actors is a recurrent MLP comprised of a GRU with a 64-dimensional hidden state, with fully connected layers on either side. We train the GRU networks on batches of 32 fully unrolled episodes (with 0-padding to account for temporal mismatch between episodes). The architecture of all shared critic networks is an MLP with 2 hidden layers with 64 units. All networks use ReLU activations for the hidden layers. All actor critic methods select discrete actions using the Gumbel-Softmax estimator~\citep{de2005tutorial} in order to turn continuous softmaxed logits into discrete one-hot actions while retaining the ability to backpropagate through the network. Actor networks and DRQNs receive the local observations of that agent as an input, appended with the last action taken by the agent, as well as a one-hot vector due to the parameter sharing. All agents use $\epsilon$-greedy action selection and we anneal $\epsilon$ from 0.5 to 0.05 over 50k training steps. The replay buffer contains the most recent 5000 episodes. All target networks are updated hard every 200 training steps. All networks are trained using Adam (except for KPG which uses RMSProp) with a learning rate of 0.0025 for the actor network and 0.0005 for the critic network (except for QMIX which uses the learning rates specified in~\citet{samvelyan2019starcraft} as they have already been tuned for SMAC).

\begin{algorithm}
   \caption{$K$-FACMAC}
   \label{A:KFACMAC}
\begin{algorithmic}
   \STATE {\bfseries Input:} Initial actor parameters $\boldsymbol{\theta}_{i}$; Initial critic parameters $\boldsymbol{\phi}_{i}$; Initial mixer parameters $\boldsymbol{\psi}$; Actor targets $\boldsymbol{\overline{\theta}_{i}}$; Critic targets $\boldsymbol{\overline{\phi}_{i}}$; Mixer target $\boldsymbol{\overline{\psi}}$; Actor optimizers $\Omega_{i}$; Critic and mixer update function $\mathcal{F}()$; recursive reasoning steps $k'$; $\forall  i\in\mathcal{I}$
   \FOR{each update step}
   \STATE Update critics and mixer using $\mathcal{F}$($\{\boldsymbol{\phi}_{t,i}\}, \boldsymbol{\psi_t}, \{\boldsymbol{\overline{\theta}_{t,i}}\}$) \COMMENT{typical FACMAC critic and mixer update}
   \STATE Save initial actor parameters $\boldsymbol{\theta}_{reset,i} \leftarrow \boldsymbol{\theta}_{t,i}$  $\forall  i\in\mathcal{I}$
   \STATE Save initial actor optimizer states $\Omega_{reset, i} \leftarrow \Omega_{t,i}$  $\forall  i\in\mathcal{I}$
   \STATE Sample $0$-level actions $a_{i}^{k=0} \sim \boldsymbol{\theta}_{t,i}$, $\forall  i\in\mathcal{I}$
   \FOR{$j=1$ {\bfseries to} $k'$}
   \STATE Sample $0$-level actions $a_{i} \sim \boldsymbol{\theta}_{t,i}$ , $\forall  i\in\mathcal{I}$
   \STATE Update actor parameters $\boldsymbol{\theta}_{t,i}^{k=j} \leftarrow \Omega_{t,i}(a_{t,i}, a_{-i}^{k=j-1}, \boldsymbol{\theta}_{t,i})$, $\forall  i\in\mathcal{I}$ \COMMENT{this implicitly updates the optimizer states}
   \IF{$j \neq k'$}
   \STATE Sample $j$-level actions $a_{i}^{k=j}\sim \boldsymbol{\theta}_{t,i}^{k=j}$ $\forall  i\in\mathcal{I}$
   \STATE Reset actor parameters $\boldsymbol{\theta}_{t,i} \leftarrow \boldsymbol{\theta}_{reset,i}$ ,  $\forall i\in\mathcal{I}$
   \STATE Reset actor optimizers $\Omega_{t,i} \leftarrow \Omega_{reset, i}$ ,  $\forall i\in\mathcal{I}$
   \ENDIF
   \ENDFOR
   \STATE Update targets $\boldsymbol{\overline{\theta}_{i}}, \boldsymbol{\overline{\phi}_{i}}, \boldsymbol{\overline{\psi}}$,  $\forall i\in\mathcal{I}$
   \STATE $\boldsymbol{\theta}_{t+1,i} \leftarrow  \boldsymbol{\theta}_{t,i}^{k=k'}$,  $\forall i\in\mathcal{I}$
   \ENDFOR
\end{algorithmic}
\end{algorithm}

\section{Additional Results}\label{appendix:tables}
\begin{table}
    \caption{Final mean performance and success rate improvement w.r.t. FACMAC for $K2$-FACMAC and baselines across MAMuJoCo and SMAC maps.}
    \label{T:exp_results}
    \vskip 0.15in
    \begin{center}
    \begin{tabular}{c|cc}
        \hline
        Algorithm  & MAMuJoCo & SMAC \\
        \hline
        FACMAC & $+0.0\%$ & $+0.0\%$\\
        COMIX/QMIX & $-53\%$ & $-26\%$\\
        MADDPG & $-238\%$ & $-85\%$\\
        POLA & $-106\%$ & $-88\%$\\
        \hline
        \textbf{K2-FACMAC} & $\mathbf{+114\%}$ & $\mathbf{+98\%}$\\
        \hline
    \end{tabular}
\end{center}
\end{table}

\begin{table}
    \caption{Mean wall clock increases to run $K$-FACMAC and $K$-MADDPG relative to nominal FACMAC and MADDPG, respectively.}
    \label{T:ablate-k}
    \vskip 0.15in
    \begin{center}
    \begin{tabular}{c|cc}
        \hline
        $K$-Level  & FACMAC & MADDPG \\
        \hline
        2 & $+28\%$ & $+27\%$\\
        3 & $+55\%$ & $+52\%$\\
        4 & $+84\%$ & $+84\%$\\
        5 & $+106\%$ & $+99\%$\\
        \hline
    \end{tabular}
\end{center}
\end{table}

\paragraph{K2-FACMAC doubles the performance of FACMAC on average.} Table~\ref{T:exp_results} shows the average performance and success rate changes for each algorithm relative to FACMAC. Note that the extreme jump for $K2$-FACMAC over FACMAC is skewed higher in these experiments due to the number of difficult environments which FACMAC never solved reasonably, such as Walker2d, 5m\_vs\_6m, Corridor, and 6h\_vs\_8z.



\end{document}